%% file: main.tex
\documentclass[10pt,twocolumn,letterpaper]{article}

\usepackage{iccv}
\usepackage{times}
\usepackage{epsfig}
\usepackage{graphicx}
\usepackage{amsmath}
\usepackage{amssymb}

\usepackage{booktabs}
\usepackage{enumitem}
\usepackage{multirow}
\usepackage{tablefootnote}
\usepackage{siunitx}
\usepackage{makecell}
\usepackage{amssymb}
\usepackage{pifont}
\usepackage{lipsum}
\usepackage{stfloats}
\usepackage{xcolor}
\usepackage{tikz}
\usepackage{wrapfig}
\usepackage{afterpage}
\usepackage{placeins}
\usepackage[abs]{overpic}
\usepackage[
  format=plain,
  labelformat=simple,
  labelsep=period,
  font=small,
  compatibility=false]{caption}
\usepackage{subcaption}
\usetikzlibrary{calc}
\usetikzlibrary{shapes}
\usetikzlibrary{arrows.meta}
\usetikzlibrary{decorations.text}
\usetikzlibrary{decorations.pathreplacing}
\usetikzlibrary{decorations,calligraphy}

\newcommand{\cmark}{\ding{51}}%
\newcommand{\xmark}{\ding{55}}%

\ExplSyntaxOn
\cs_new:Npn \expandableinput #1
  { \use:c { @@input } { \file_full_name:n {#1} } }
\ExplSyntaxOff

\input{definitions.tex}

\usepackage[pagebackref=true,breaklinks=true,letterpaper=true,colorlinks,bookmarks=false]{hyperref}
\hypersetup{colorlinks,linkcolor={blue},citecolor={blue}}

\definecolor{myblue}{HTML}{457b9d}
\definecolor{myred}{HTML}{E63946}
\definecolor{mygreen}{HTML}{40916c}
\definecolor{myyellow}{HTML}{ffbe0b}
\definecolor{myorange}{RGB}{247,214,157}
\definecolor{mygray}{RGB}{211,211,211}
\definecolor{mypurple}{RGB}{212,164,217}
\definecolor{gray3}{gray}{0.5}
\definecolor{gray2}{gray}{0.7}
\definecolor{gray1}{gray}{0.85}

\definecolor{burstred}{RGB}{235,50,35}
\definecolor{burstgreen}{RGB}{140,206,89}
\definecolor{burstblue}{RGB}{75,174,234}

\usepackage[capitalize]{cleveref}
\crefname{section}{Sec.}{Secs.}
\Crefname{section}{Section}{Sections}
\Crefname{table}{Table}{Tables}
\crefname{table}{Tab.}{Tabs.}

\iccvfinalcopy % *** Uncomment this line for the final submission

\def\iccvPaperID{10091} % *** Enter the ICCV Paper ID here

\ificcvfinal\fi

\begin{document}

%%%%%%%%% TITLE
\title{Towards Real-World Focus Stacking with Deep Learning}

\author{Alexandre Araujo \qquad Jean Ponce \qquad Julien Mairal\\[0.1cm]
INRIA, Ecole Normale Supérieure, CNRS, PSL University, Paris, France
}

% \maketitle

\twocolumn[{%
  \small
  \maketitle
  \centering
  \vspace{-0.4cm}
  \begin{tabular}{cc}
    \includegraphics[width=0.47\textwidth]{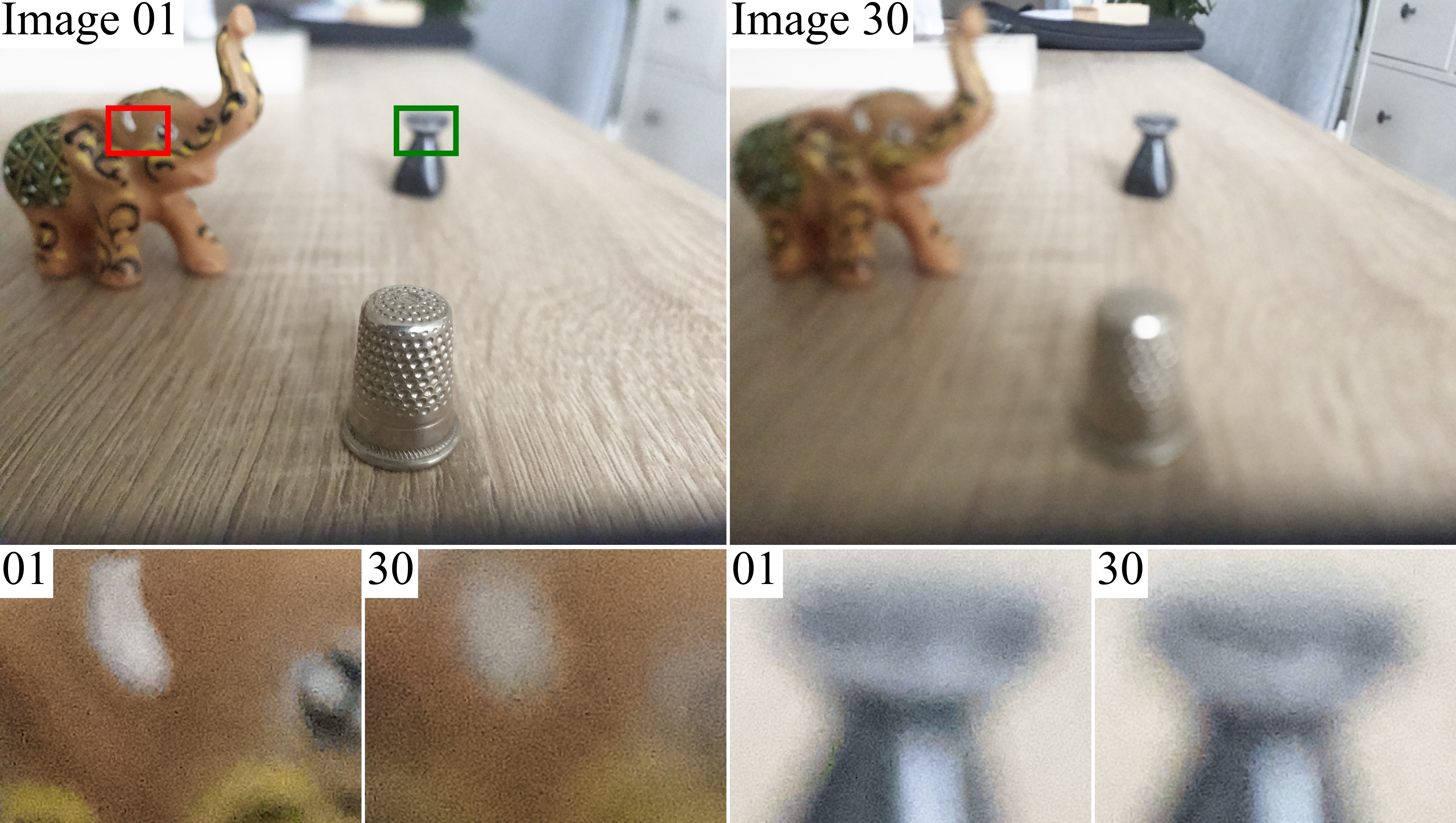} & 
    \includegraphics[width=0.47\textwidth]{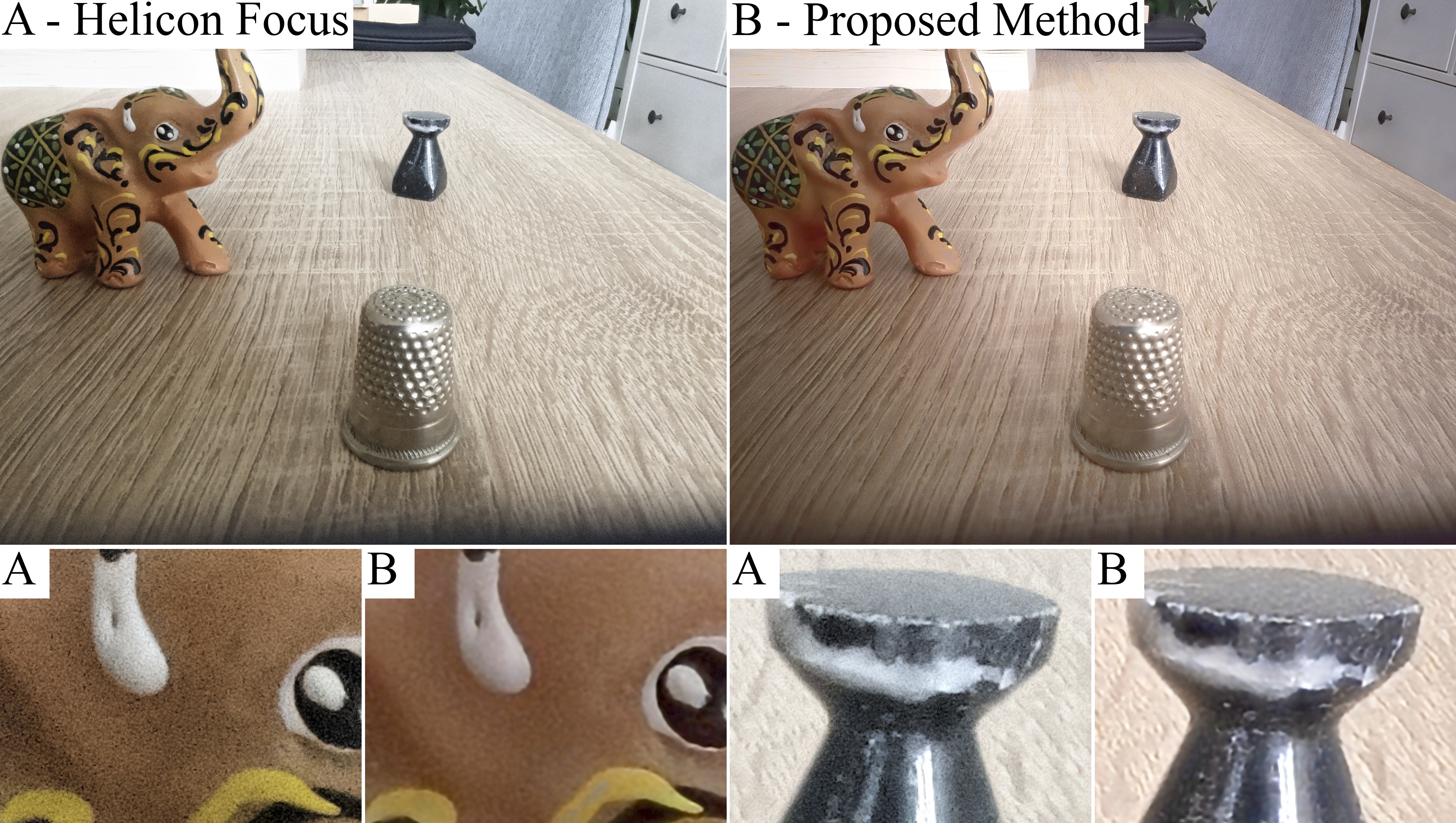} \\
  \end{tabular}
  \vspace{-0.3cm}
  \captionof{figure}{Left: the first and last images of a burst with 30 focus-bracketed raw images taken by an iPhone 12 at 1000 ISO. Right: the results obtained by a state-of-the-art commercial software for focus stacking, HeliconFocus (A), and the deep learning approach proposed in this paper (B). In this short-exposure, high-ISO setting aimed at minimizing motion blur, our solution recovers the same level of detail as HeliconFocus, but achieves a significant level of noise reduction. Best seen by zooming on a computer screen.}
  \vspace*{0.3cm}
  \label{fig:teaser}
}]

\ificcvfinal\thispagestyle{empty}\fi

%%%%%%%%% ABSTRACT
\begin{abstract}
\vspace{-0.1cm}
Focus stacking is widely used in micro, macro, and landscape photography to reconstruct all-in-focus images from multiple frames obtained with focus bracketing, that is, with shallow depth of field and different focus planes. Existing deep learning approaches to the underlying multi-focus image fusion problem have limited applicability to real-world imagery since they are designed for very short image sequences (two to four images), and are typically trained on small, low-resolution datasets either acquired by light-field cameras or generated synthetically. We introduce a new dataset consisting of 94 high-resolution bursts of raw images with focus bracketing, with pseudo ground truth computed from the data using state-of-the-art commercial software. This dataset is used to train the first deep learning algorithm for focus stacking capable of handling bursts of sufficient length for real-world applications. Qualitative experiments demonstrate that it is on par with existing commercial solutions in the long-burst, realistic regime while being significantly more tolerant to noise.
The code and dataset are available at \url{https://github.com/araujoalexandre/FocusStackingDataset}.
\end{abstract}

%%%%%%%%% BODY TEXT

\begin{table*}[ht]
  \centering
  {\footnotesize
  \caption{This table compares existing datasets for multi-focus image fusion. Most datasets are synthetically generated either by simulated blur or via light-field camera. These datasets are too small to be used as training sets for supervised methods and feature very short (2 to 4) sequences unsuitable for real-world focus stacking applications. Our Large-Scale Focus Dataset (LSFD) dataset is composed of 94 high-resolution raw bursts of 30 images. The ground truth is generated via HeliconFocus, a state-of-the art focus stacking software. For training, we extract 1200 crops on each high-resolution burst leading to more than 11Ok $128\times128$ bursts of 30 images. We don't take into account data augmentation (rotation, flip, etc.) in the number of bursts.}
  \label{table:datasets}%
    \begin{tabular}{lccccccr}
    \toprule
    \textbf{Name} & \textbf{Method} & \makecell{\textbf{Publicly} \\ \textbf{Available}} & \textbf{Resolution} & \makecell{\textbf{Raw} \\ \textbf{Available}} & \makecell{\textbf{Number of} \\ \textbf{Bursts}} & \makecell{\textbf{Length of} \\ \textbf{Burst}} & \makecell{\textbf{Total number} \\ \textbf{of pixels}} \\
    \midrule
    % Savic et al. (2012) \cite{savic2012multifocus} &  & Various & \xmark & 27 & 2 & 54 \\
    MLCNN \cite{zhao2018multi}      & Synthetic & \xmark & $41\times41$ & \xmark & \num{291} & 2 & 0.9M\ \ \ \ \ \ \  \\
    CNN \cite{liu2017multi}         & Synthetic & \xmark & $16\times16$ & \xmark & \num{10000} & 2 & 5.1M\ \ \ \ \ \ \  \\
    BA-Fusion \cite{ma2019boundary} & Synthetic & \xmark & $16\times16$   & \xmark & \num{18900} & 2 & 9.6M\ \ \ \ \ \ \  \\
    Lytro \cite{nejati2015multi}    & Light field camera & \cmark & $520\times520$ & \xmark & 20 & 2 & 10.8M\ \ \ \ \ \ \   \\
    MFFW \cite{xu2020mffw}          & Hand crafted ground truth & \cmark & Various & \xmark & 13 & 2 & 21M\ \ \ \ \ \ \  \\
    MF \cite{aymaz2020multi}        & Synthetic & \cmark & $512\times512$ & \xmark & 33 & 4 & 34M\ \ \ \ \ \ \  \\
    MFI-WHU \cite{zhang2021mff}     & Synthetic & \cmark & Various & \xmark & 120 & 2 & 63M\ \ \ \ \ \ \ \\
    Real-MFF \cite{zhang2020real}   & Light field camera & \xmark & $625\times433$ & \xmark & 710 & 2 & 384M\ \ \ \ \ \ \  \\
    FuseGan \cite{guo2019fusegan}   & Synthetic & \xmark & $320\times480$ & \xmark & \num{5850}    & 2 & 1.7B\ \ \ \ \ \ \  \\
    \midrule
    \multirow{2}{*}{\textbf{LSFD (ours)}} & \multirow{2}{*}{\makecell{Real-world focus burst with \\ pseudo ground truth via HeliconFocus}} & \multirow{2}{*}{\cmark} & \multirow{2}{*}{$5184\times3888$} & \multirow{2}{*}{\cmark} & \multirow{2}{*}{94} & \multirow{2}{*}{30} & \multirow{2}{*}{62.7B\ \ \ \ \ \ \ } \\
    & & & & & & \\
    \bottomrule
    \end{tabular}%
  }
  % \vspace{-0.3cm}
\end{table*}%

%%%%%%%%% BODY TEXT
\vspace{-0.2cm}
\section{Introduction}
\label{section:introduction}

Focus stacking is a well-known technique using multiple images of the same scene with a low depth of field and different focusing planes in order to obtain an all-in-focus image with a much larger depth of field.
This method has a wide range of applications including macro~\cite{cremona2014extreme,dalrymple2014focus,marziali2017focus}, and micro photography~\cite{dendere2011image,goldsmith2000deep,hovden2011extended,sigdel2015focusall,valdecasas2001extended}, etc.
In these settings, the camera is known to have a low depth of field, and capturing the entire object fully in-focus is physically impossible.

Focus stacking has been studied for decades with ``classical'' approaches (\ie, handcrafted without learning), first with pixel-based methods~\cite{pieper1983image,sugimoto1985digital}, then with pixel neighborhood information to reconstruct the full image based on detected sharp pixels~\cite{goldsmith2000deep,tympel1996new}.
Multi-resolution approaches \cite{forster2004complex,li1995multisensor} such as wavelets \cite{mallat1989theory}, have also shown great success.
More recently, supervised and unsupervised learning approaches using deep learning have been proposed~\cite{li2020drpl,liu2017multi,ma2021sesf,tang2018pixel,wang2022multi,xiao2021dtmnet,xiao2020global,xu2020u2fusion,zhang2021mff,zhang2020rethinking,zhang2020ifcnn}.
However, given the physical properties of camera lenses, constructing a real-world dataset with focus bracketing and sharp images as ground truth is a very challenging task.
To address this issue, semi-synthetic datasets have been generated from all-in-focus images, either by simulating bursts with synthetic blur or with light-field cameras~\cite{aymaz2020multi,guo2019fusegan,liu2017multi,ma2019boundary,nejati2015multi,xu2020mffw,zhang2021mff,zhang2020real}.
To the best of our knowledge, such datasets are unfortunately quite unrealistic.
Indeed, most datasets with simulated blur have been designed with only two images and low resolution, which makes them unsuitable for high-resolution micro and macro photography where focus bursts are usually large (\ie, more than 20 images). 
Furthermore, generating realistic simulated blur from real in-focus images is very hard since this requires both an accurate physical model of the blurring process (thin lens models are a very coarse approximation of real lenses, in particular in the macrophoto domain) and depth information which is often lacking \cite{zhang2020real}.
Datasets based on images from light-field cameras are more realistic, but resolution remains limited~\cite{zhang2020real}.

Most existing approaches using deep learning focus mainly on image fusion with very short image sequences (two to four images)  with a sharp foreground and a blurry background or vice versa.
This setting is very useful to get an all-in-focus image of a subject (\eg, person) with a background (\eg, landscape).
% While in the setting of multi-focus image fusion, a small number of images is usually sufficient to obtain satisfying results, 
True focus stacking, on the other hand, is usually applied on bursts with much lower depth-of-field and requires several dozens images (sometimes hundreds) to recover an all-in-focus image in the context of macro and micro photography.

In this paper, we propose a different dataset generation process along with a deep learning method that is able to demosaic and focus stack high-resolution bursts of focus-bracketed raw images.
Instead of generating bursts from an all-in-focus image and simulated blur, we generate a dataset based on real-world raw bursts taken with focus bracketing and a pseudo ground truth computed with a state-of-the-art commercial software, HeliconFocus~\cite{kozub2008helicon}.
Even though using a pseudo ground truth for learning may appear to be a strong limitation at first sight, this approach offers several advantages over previous work.
First, we generate high-resolution bursts in good conditions with little noise, which allows us to have good quality images.
The length of our bursts is large (in our case, 30 images by burst), which is more in line with focus stacking performed by photographers.
Second, during training, we benefit from true blur which is very hard to simulate, and we can afford to inject realistic noise, which allows us to be more robust than the original method used to generate the pseudo ground truth.
Finally, high-resolution bursts can be split into smaller crops leading to a large dataset for supervised training.
To the best of our knowledge, our Large-Scale Focus Dataset (LSFD) is over 35 times larger than the largest dataset currently available for focus stacking by number of pixels.
Our main contributions can be summarized as follows:
\begin{itemize}[leftmargin=10pt,parsep=0pt,topsep=0pt,itemsep=2pt]
  \item We construct and release a new large-scale dataset composed of 94 bursts with focus bracketing with 30 high-resolution raw images for each burst and pseudo ground truth computed with state-of-the-art software.
  \item We propose a deep learning architecture for focus stacking directly from raw bursts, extending the work of \cite{luo2021ebsr} to this task. Furthermore, we inject realistic noise during training in order to make our approach more robust to noise than the method used to compute our ground truth.
  \item We demonstrate the effectiveness of our approach with extensive quantitative and qualitative comparisons.
\end{itemize}

\section{Related Work}
\label{section:related_work}

\subsection{Multi-Focus Image Fusion Datasets}
\label{section:related_work-datasets}

There exist two types of focus stacking datasets: those based on bursts with simulated blur and those based on bursts constructed from light-field cameras.
These two kinds of datasets offer some trade-offs.
Indeed, it is easier to scale the number of bursts using simulated blur than with light-field cameras.
Table~\ref{table:datasets} describes previous datasets for focus stacking and highlights their characteristics.
We observe that datasets with a large number of bursts and total images are not publicly available.
Below, we describe some methods used to construct these datasets.

\vspace{-0.2cm}

\paragraph{Datasets with simulated blur.}
The first dataset generated with simulated blur was proposed by \cite{liu2017multi}.
They used Gaussian filters on a portion of the images to simulate shallow depth-of-field.
The dataset was generated from the validation set of ImageNet dataset \cite{deng2009imagenet}.
Given that kernel filters are spatially invariant, the blur is not realistic since depth variation is not taken into account.
In order to take depth information into account, several works~\cite{aymaz2020multi,guo2019fusegan,ma2019boundary,zhang2021mff,zhao2018multi} use segmentation datasets~\cite{lin2014microsoft} and extract the foreground and background of the images, then apply Gaussian blur on these segmented portions to construct focus bursts with two images.
In a similar line of work, \cite{xu2020mffw} proposes a new dataset with real defocus bursts collected online and constructs the ground truth by manually annotating sharp portions of the images. 
While this approach can be interesting given that bursts are not synthetically generated, the dataset is very small and therefore unsuitable for supervised learning.
We argue that generating synthetic but realistic focus-bracketed bursts with a very shallow depth of field is a very challenging task and remains an open problem.

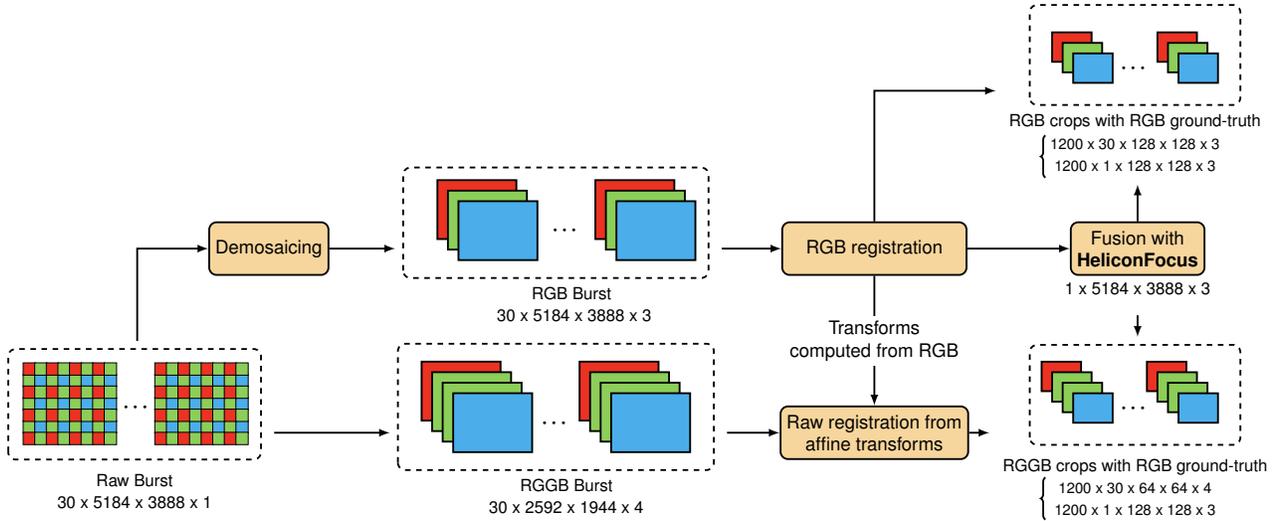
\begin{figure*}[t]
  \centering
  \scalebox{0.70}{%
    \input{figures/datasets}
  }%
  \caption{This figure describes the creation of the raw and RGB datasets for supervised learning from raw bursts taken with the Panasonic Lumix DC-GX9. First, the raw images are represented as a Bayer pattern, then the raw images are converted to RGB bursts through a process called demosaicing. Second, each image of the RGB burst are registered with each other with the enhanced correlation coefficient (ECC) maximization algorithm~\cite{evangelidis2008parametric}. The affine transforms are saved and applied to the RGGB burst. From the aligned RGB burst we compute our pseudo ground truth with HeliconFocus~\cite{kozub2008helicon}. Finally, the high-resolution burst and the associated pseudo ground truth are split into small $128 \times 128$ crops to make them more suitable for training in mini-batches.}
  \label{figure:dataset_pipeline}
\end{figure*}

\paragraph{Datasets with light-field cameras.}

Unlike conventional cameras, light-field cameras are able to capture information about the entire {\em light field} emanating from a scene.
Images are then generated afterward based on the information collected.
\cite{ng2005fourier} proposed a method based on the {\em Fourier slice photography theorem} to compute photographs focused at different depths from a single light field.
This method makes it possible to generate a simulated depth of field from the light field information.
Although datasets from light-field cameras \cite{nejati2015multi,zhang2020real} have the potential to provide bursts with realistic blur and ground truth, current datasets are quite small in size and resolution, making them unsuitable for supervised learning as well.

\subsection{Focus Stacking Methods}

\paragraph{Classical Methods.}

One of the first methods to combine several images to obtain a single image with extended depth-of-field dates back several decades~\cite{adelson1987depth,itoh1989digitized,pieper1983image,sugimoto1985digital}.
These methods were pixel-based with maximum or minimum selection to select the in-focus frame.
Although simple, these methods suffer from pixel discontinuity as local information around the pixel was not used. 
Later, works based on pixel neighborhood information (\eg, the variance around the pixel) have been used to better detect in-focus pixels in the burst \cite{goldsmith2000deep,tympel1996new}.
However, great improvement has been made with multi-resolution analysis.
This approach leverages the assumption that in-focus parts of the image contain high-frequency components.
One of the first methods to demonstrate the success of this approach was introduced in~\cite{burt1992gradient,burt1993enhanced} and based on applying a Laplacian operator on each image of the stack in order to determine the best in-focus pixel value and merging all the images based on the maximum value of this operator on each image.
Since then, these multi-resolution analysis methods, mostly based on wavelets, have been extensively studied and are, to the best of our knowledge, the de facto approaches to focus stacking, especially in microphotography~\cite{forster2004complex,hill2002image,lewis2007pixel,li1995multisensor,li2008multifocus,li2011performance,shi2005wavelet,valdecasas2001extended}.
More recently, a multitude of different works have tried to improve upon the state of the art using for example a gradient-based approach~\cite{zhang2020rethinking,zhou2014multi}, a frequency-based approach based on the discrete cosine transform~\cite{cao2014multi}, a quad-tree decomposition~\cite{bai2015quadtree,de2013multi,wang2022multi}, image matting~\cite{chen2021multi,li2013image,ma2020alpha}, guided filtering~\cite{li2013imagefusion,qiu2019guided,singh2020construction}, sparse representations~\cite{liu2016image,liu2015general} or different focus measures~\cite{ma2017multi,sigdel2015focusall,zhang2017boundary}.

The success of these methods has made them popular with commercial software specializing in focus stacking.
For example, without giving much detail, HeliconFocus~\cite{kozub2008helicon} offers three different methods.
The first method is a pixel-wise method based on a measurement of the contrast of each pixel.
The second one takes neighborhood information based on the variance of surrounding pixels.
And finally, the third method adopts a pyramidal approach by splitting the image signals into high and low frequencies.

\begin{table}[t]
  \centering
  \caption{This table describes the breakdown of bursts taken with the Leica and Olympus lenses as well as the split between train and test set for training and evaluation. Figure~\ref{figure:burst-dataset} in the Appendix presents several bursts for each camera lens.}
  {\normalsize
  \begin{tabular}{lcc}
  \toprule
    \textbf{Lenses} & \textbf{Train Set} & \textbf{Test Set} \\
  \midrule
  Leica 25mm/F1.4 II & 47 & 5 \\
  Olympus M.60mm F2.8 Macro & 37 & 5 \\
  \midrule
  \textbf{Total} & \textbf{84} & \textbf{10} \\
  \bottomrule
  \end{tabular}%
  }
  \label{table:split_dataset}%
  \vspace{-0.3cm}
\end{table}%

\begin{figure}[t]
  \centering
  \includegraphics[width=0.47\textwidth]{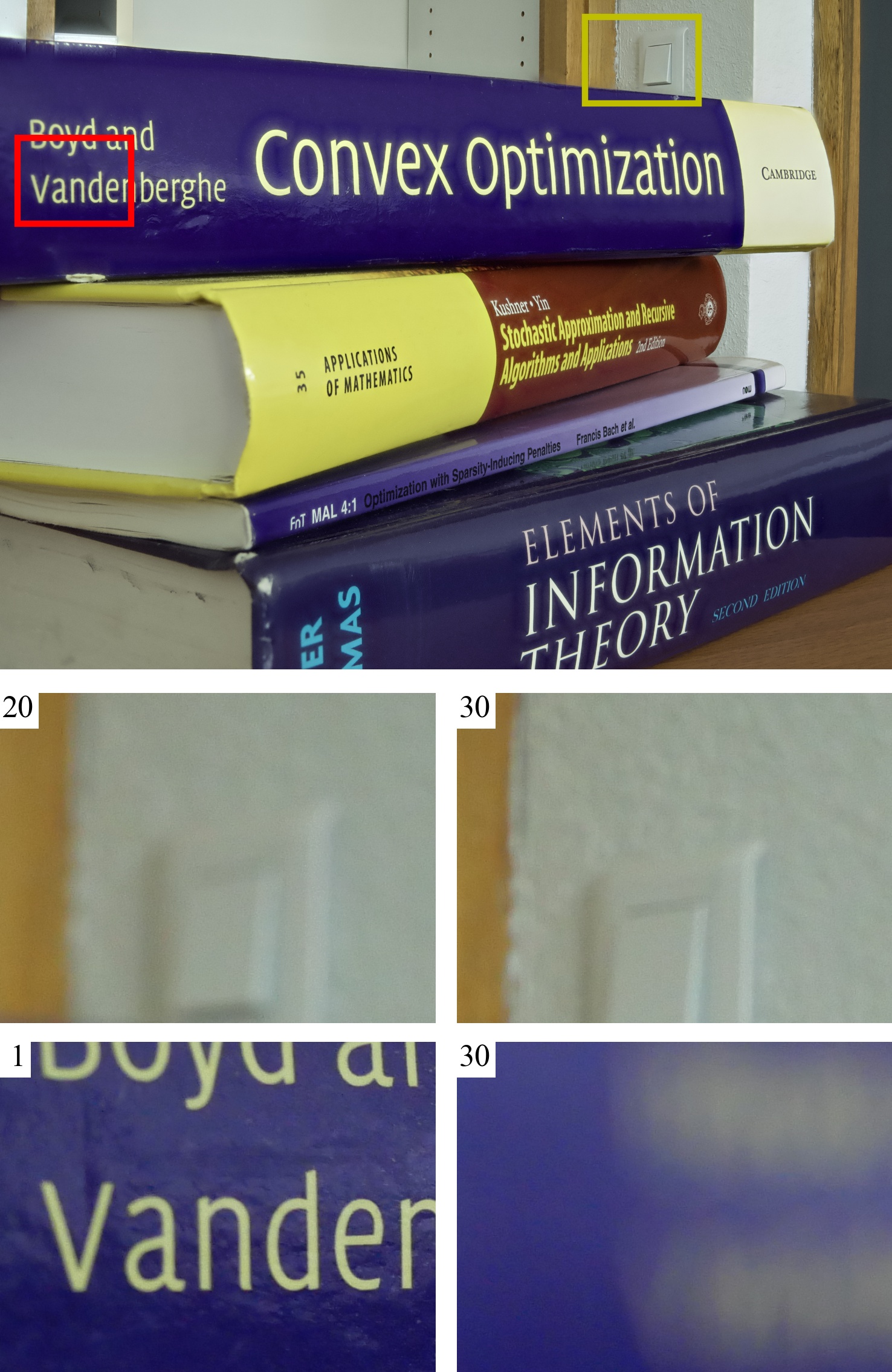}
  \caption{This figure shows misalignment due to changing magnification during the focus bracketing.}
  \label{figure:misalignment}
  \vspace{-0.3cm}
\end{figure}

\vspace{-0.2cm}
\paragraph{Deep learning methods.}
Deep learning approaches to multi-focus images fusion include supervised~\cite{li2020drpl,liu2017multi,tang2018pixel,xiao2021dtmnet,xiao2020global,zhang2020ifcnn} as well as unsupervised ones~\cite{ma2021sesf,xu2020u2fusion,zhang2021mff}.
They use large training datasets with simulated blur from all-in-focus images (Section~\ref{section:related_work-datasets}).
Conventional methods essentially consist of two steps where first a level measurement is performed on either the spatial domain or in a transform domain like wavelets and then a fusion is performed based on this measurement.
Deep learning approaches learn the defocus measure (whether a pixel is in focus or not) and the fusion rule simultaneously.
It should be noted that these approaches using deep learning focus mainly on image fusion with at most four images where only the foreground is sharp and the background is blurred and vice versa.
This practice is very useful when an image has a subject in the foreground and the background is blurry.
To the best of our knowledge, only one approach~\cite{xiao2022general} proposed a deep learning architecture that can take more than two images, however, they experiment with multi-focus image fusion datasets with 2 and 3 images by burst.

A lot of methods published in the past 10 years among ``classical'' or deep learning approaches evaluate their approach with a multitude of metrics (\cite{zhang2021deep} presents nineteen metrics used in recent literature) and several datasets~\cite{nejati2015multi,xu2020mffw,zhang2021mff} which do not have a ground truth.
Therefore, it is very difficult to properly characterize the real improvement of their approach compared with wavelet-based approaches.

\begin{figure*}[t]
  \centering
  \scalebox{0.75}{%
    \input{figures/architecture}
  }%
  \caption{The figure presents our FocusDeep architecture inspired by the architecture proposed by~\cite{luo2021ebsr} in the context of Super-Resolution from raw bursts and adapted for focus stacking. The model is designed in three blocks, the first one consisting of a feature extractor and an alignment layer. The second block aims at merging all the frames into one with a convolutional layer. Finally, the architecture ends with a reconstruction layer and a long-range skip-connection.}
  \label{figure:deep_architecture}
\end{figure*}
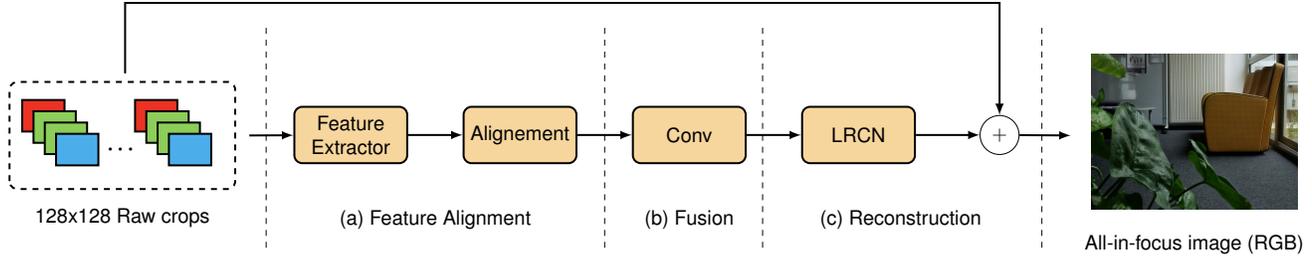

\section{Datasets \& Methods}

This section presents our Large-Scale Focus Dataset (LSFD), the approach used to obtain a pseudo ground truth and we describe the deep learning architectures used for training.
To the best of our knowledge, this dataset is the largest ever devised in the context of focus stacking and multi-focus image fusion and the only one with raw bursts.

\subsection{Methodology}

The dataset was been devised with a Panasonic Lumix DC-GX9\footnote{\url{https://www.panasonic.com/uk/consumer/cameras-camcorders/lumix-mirrorless-cameras/lumix-g-cameras/dc-gx9.html}} with two different camera lenses: a Leica 25mm/F1.4 II and an Olympus M.60mm F2.8 Macro.
Although the Olympus Macro lens is more in line with the objective of performing focus stacking on macro photography, we noticed that the depth-of-field is too shallow to gather all parts of the scene into focus with a burst of 30 frames.
Therefore, we decided to design the dataset with a classical lens as well as a macro lens. 
Table~\ref{table:split_dataset} describes the breakdown of bursts taken with each lens as well as the split between the train and test set.
We have taken 52 and 42 bursts of 30 raw images with each of the Leica and Olympus lenses respectively.
We provide in Figure~\ref{figure:burst-dataset} in the Appendix several bursts for each camera lens.
To capture each burst, we used the focus bracketing function of the Panasonic camera which automatically shoots a series of images at a range of different focus positions.
The main difficulty in creating our dataset was to minimize camera and scene motion to simplify alignment, which is difficult in focus stacking because corresponding regions may look different due to blur and magnification that changes with focus, particularly in macro settings. 
In order to avoid (as much as possible) misalignment issues between shots, we use a tripod as well as a 10 seconds delay between the press of the button and the start of the shots to let the camera stabilize.
In addition, a qualitative review of all bursts was performed, and a total of 34 bursts were discarded due to scene motion.

\begin{figure*}[t]
  \centering
  \begin{subfigure}[t]{1\textwidth}
    \centering
    \includegraphics[width=0.98\textwidth]{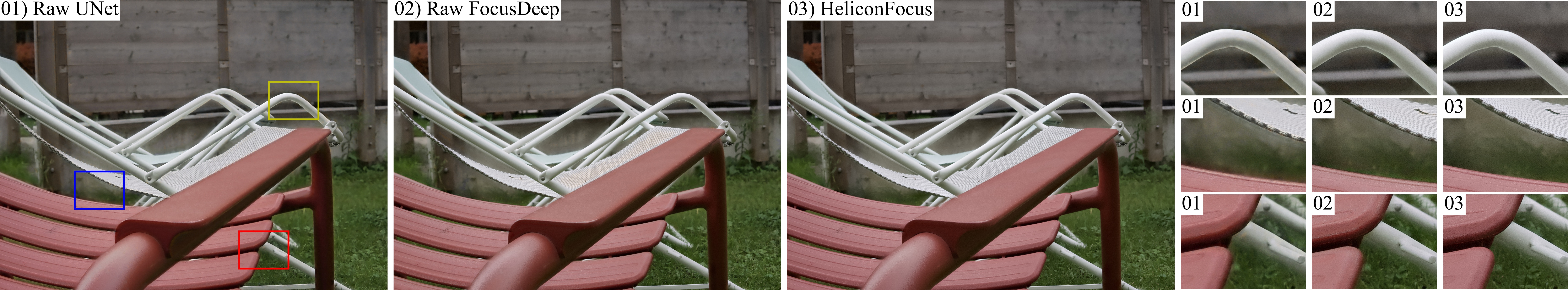}
  \end{subfigure}%
  \vspace{0.1cm}
  \begin{subfigure}[t]{1\textwidth}
    \centering
    \includegraphics[width=0.98\textwidth]{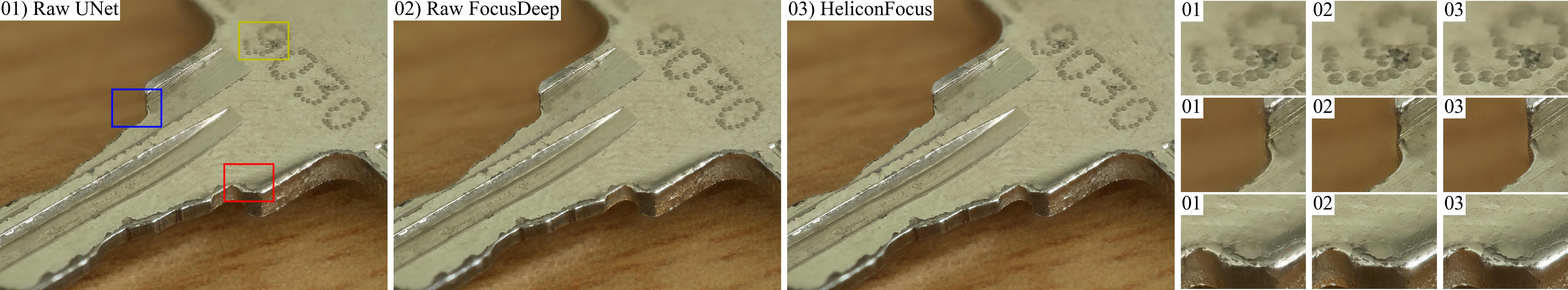}
  \end{subfigure}%
  \caption{Results of the UNet and FocusDeep architecture with raw bursts from our test dataset. Top row: the result of a burst taken with the Leica lens. Bottom row: the result of a burst taken with the Olympus lens. We observe some small artifacts with the UNet architecture results confirming the lower quantitative results. Additional results on the training dataset are shown in Figure~\ref{figure:results_deep_learning_train}.}
  \label{figure:results_deep_learning_test}
  \vspace{-0.2cm}
\end{figure*}

\subsection{Processing of Raw Bursts}

After collecting a set of raw bursts, we designed a pipeline to process them and build two datasets suitable for supervised learning. 
Figure~\ref{figure:dataset_pipeline} shows an overview of this pipeline.
First, we convert our raw bursts to RGB by using an off-the-shelf demosaicing algorithm, we use the ``scriptable image processing system'' library\footnote{\url{https://www.unix.com/man-page/osx/1/sips/}} available by default on MacOS.
Following the demosaicing, we aligned the images of each RGB burst with the enhanced correlation coefficient (ECC) Maximization algorithm~\cite{evangelidis2008parametric} available in the OpenCV library~\cite{bradski2000opencv} with carefully tuned hyperparameters to finely align successive and thus minimally different image pairs in each burst.
Correct registration of the frames on each burst is crucial to avoid the magnification problem of images with different focus.
Indeed, between the first and last frame of each burst a magnification effect is clearly visible and should be corrected.
Figure~\ref{figure:misalignment} shows misalignment between frames due to the focus effect.
To register images, we leverage the order of images in the burst by aligning each image with the previous one in the burst.
After aligning RGB images, we save each affine transforms, scaled them, and apply them on raw (RGGB) images.
Then, we compute our pseudo ground truth from our registered RGB bursts with HeliconFocus~\cite{kozub2008helicon} which is a state-of-the-art commercial software specifically developed for focus stacking.
We use HeliconFocus' pyramid stacking method which provides the best results.
Figure~\ref{figure:burst-heliconfocus} in the Appendix shows the result of HeliconFocus on four bursts.
Finally, we construct our RGB and raw datasets by splitting each high-resolution RGB and raw burst with their associated RGB ground truth into $128\times128$ smaller bursts more suitable for training on GPUs.

\subsection{Deep Learning Architectures}

We use two deep learning architectures for training on our dataset, both architectures are used on the RGB and raw datasets.
First, we use a plain UNet architecture \cite{ronneberger2015u} as a baseline.
Second, we propose a new architecture, called FocusDeep, inspired by the architecture proposed by~\cite{luo2021ebsr} in the context of Super-Resolution from raw bursts and adapted to focus stacking.
This architecture consists of three modules, the first consisting of a feature extractor and an alignment module.
The alignment module, which is employed to align crops with each other during training, is a variant of the feature-enhanced pyramid cascading and deformable convolution proposed by~\cite{luo2021ebsr} and inspired by~\cite{zhu2019deformable,wang2019edvr}.
The major difference is that instead of aligning a specific reference frame (\ie, \cite{luo2021ebsr} use the first frame as reference), we align each frame relative to the previous one.
This makes more sense in the context of focus stacking as the first and last frames are difficult to align correctly due to the large blur between them.
Aligning crops during training was necessary because we observed that despite proper alignment of high-resolution images, some small shifts remained at the local level (sub-pixel alignment) on the 128x128 crops.
The second block aims at merging all the frames into one using a convolutional layer.
\cite{luo2021ebsr} use a layer called {\em Cross local Fusion} which uses non-local relations between the base frame and neighboring frames to perform their fusion.
However, to reduce the computational complexity of this module, they use the first image of the burst as a reference. 
We found this approach not suitable for focus stacking, as it induces a significant bias towards the first image of the burst.
Therefore, we replaced this layer with a convolutional layer instead.
Finally, we reconstruct the RGB image with a long-range concatenation network (LRCN), inspired by~\cite{yu2018wide}, which is composed of long-range concatenation groups where each is composed of residual blocks with wide activation (WARB).
This module offers better feature representation capabilities for the reconstruction as shown by~\cite{luo2021ebsr}.
An overview of the proposed FocusDeep architecture is shown in Figure~\ref{figure:deep_architecture}.

\section{Experiments}
\label{section:experiments}

\begin{table}[t]
  \centering
  \caption{This table presents the results in terms of the PSNR and SSIM metrics on several methods and models using HeliconFocus as ground truth. First, the results with Laplace and Wavelets models are presented as a baseline. 
Then, the results for our two deep learning architectures on our two datasets are presented. We observe that deep learning models outperform both Laplace and Wavelets models. We also observe that models which perform only focus stacking outperforms the ones which perform joint demosaic and focus stacking together.} 
  \label{table:results}
  {\small
    \begin{tabular}{llcccc}
    \toprule
    \multicolumn{2}{c}{\multirow{2}[4]{*}{\textbf{Models}}} & \multicolumn{2}{c}{\textbf{PSNR} $\uparrow$} & \multicolumn{2}{c}{\textbf{SSIM} $\uparrow$} \\
    \cmidrule{3-6}   
    & & Train & Test & Train & Test \\
    \midrule
    \multirow{2}{*}{RGB} & Laplace  & 25.92 & 26.88 & 0.659 & 0.712 \\
     & Wavelets & 26.27 & 27.65 & 0.797 & 0.828 \\
    \midrule
    \multirow{2}{*}{RGB} & Unet  & 35.13 & 36.29 & 0.899 & 0.922 \\
    & FocusDeep & 36.99 & 38.47 & 0.942 & 0.965 \\
    \midrule
    \midrule
    \multirow{2}{*}{Raw} & Unet  & 30.01 & 30.21 & 0.759 & 0.767 \\
    & FocusDeep & 33.47 & 32.89 & 0.882 & 0.898 \\
    \bottomrule
    \end{tabular}%
  }
  \vspace{-0.2cm}
\end{table}%

\begin{figure}[t]
  \centering
  \vspace{-0.65cm}
  \includegraphics[width=0.46\textwidth]{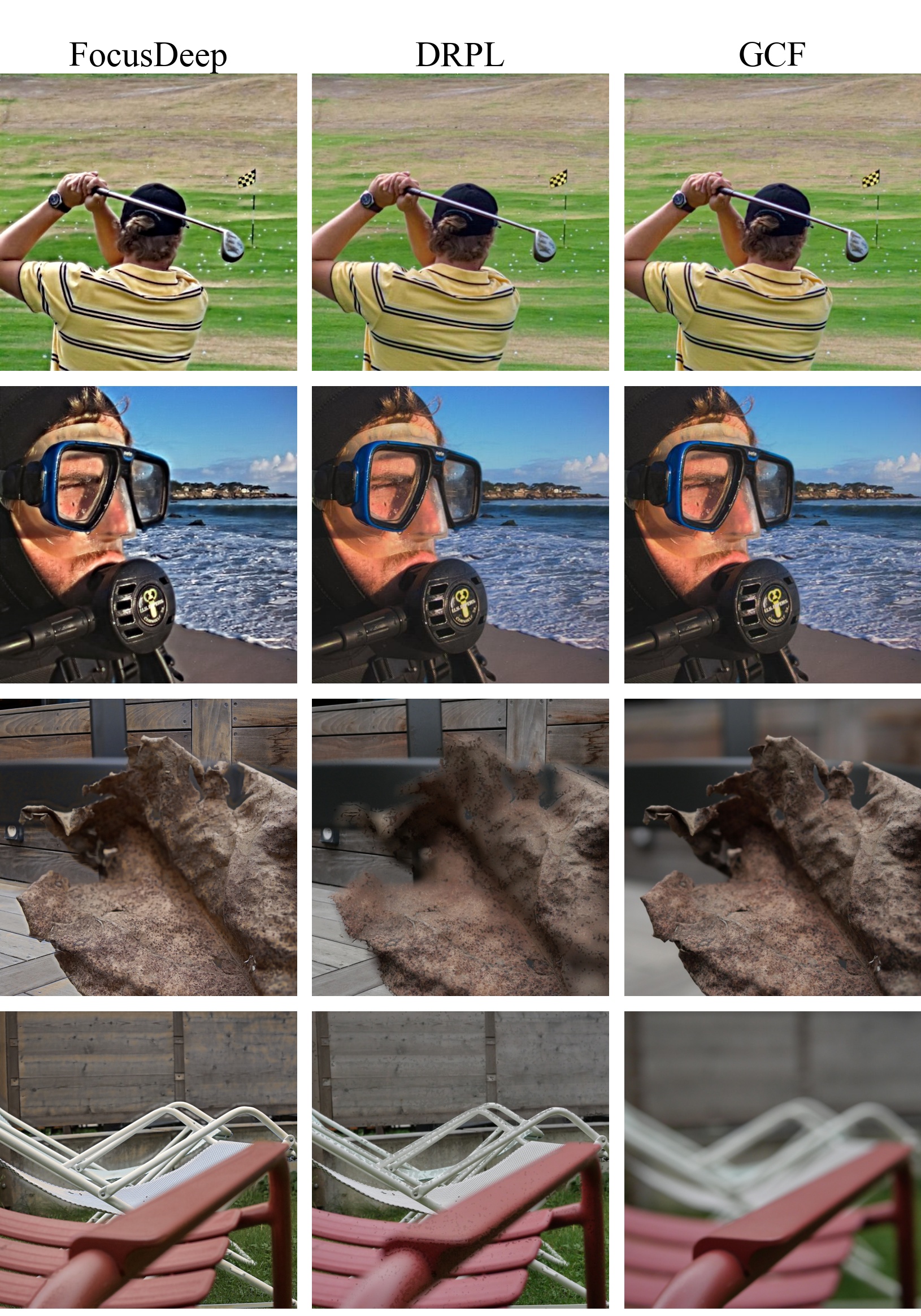}
  \caption{This figure compares our FocusDeep architecture on the Lytro dataset \cite{nejati2015multi} against the DRPL~\cite{li2020drpl} and GCF~\cite{xu2020deep} deep learning methods. The top two rows show the results on the Lytro dataset and the bottom two show the results on our RGB dataset. We observe that our method works equally well as DRPL and GCF. However, these methods do not generalize to our dataset.}
  \label{figure:others_results}
  \vspace{-0.1cm}
\end{figure}

\paragraph{Quantitative and Qualitative Evaluation.}

We compare the outputs of our models to those of HeliconFocus to ensure that we have successfully learned to perform focus stacking from RGB and raw bursts.
Table~\ref{table:results} presents our quantitative results with PSNR and SSIM metrics \cite{wang2004image}.
We have also included, for comparison, the quantitative results of a simple focus stacking performed with a Laplacian transform and wavelets transform. The wavelets model follows the work of~\cite{forster2004complex} which uses complex Daubechies wavelets.
We observe from Table~\ref{table:results} that our deep learning architectures outperform the Laplacian and wavelets models by a high margin. 
Since performing joint demosaic and focus stacking together is a more difficult task, we observe a slight decrease in the score of our models trained on the raw dataset.

Figure~\ref{figure:results_deep_learning_test} shows qualitative results of our UNet and FocusDeep architectures with the raw dataset on our test dataset.
We observe in the images some small artifacts on the results of the UNet architecture confirming the lower score.
We also present some qualitative results on the Lytro dataset and compare our results to two recent deep learning approaches: DRPL~\cite{li2020drpl} and GCF~\cite{xu2020deep}. 
We evaluate these approaches on our dataset by taking the first and last frames of the burst.
The FocusDeep result in Figure~\ref{figure:others_results} also uses the first and last frames of the burst for a fair comparison.
Figure~\ref{figure:others_results} shows the results on the Lytro dataset (top two rows), we observe that our method (which is not trained on Lytro) generalizes and performs equally well as DRPL and GCF.
On the other hand, we apply DRPL and GCF models on two bursts of our dataset and notice that these approaches do not generalize well (as expected) since our images do not have the binary foreground/background nature these approaches are designed for.
One can also note that quantitative comparison is unfortunately impossible on the Lytro dataset because this dataset does not have any ground truth.

\begin{figure}[t]
  \centering
  \begin{subfigure}[t]{0.48\textwidth}
    \centering
    \includegraphics[width=\textwidth]{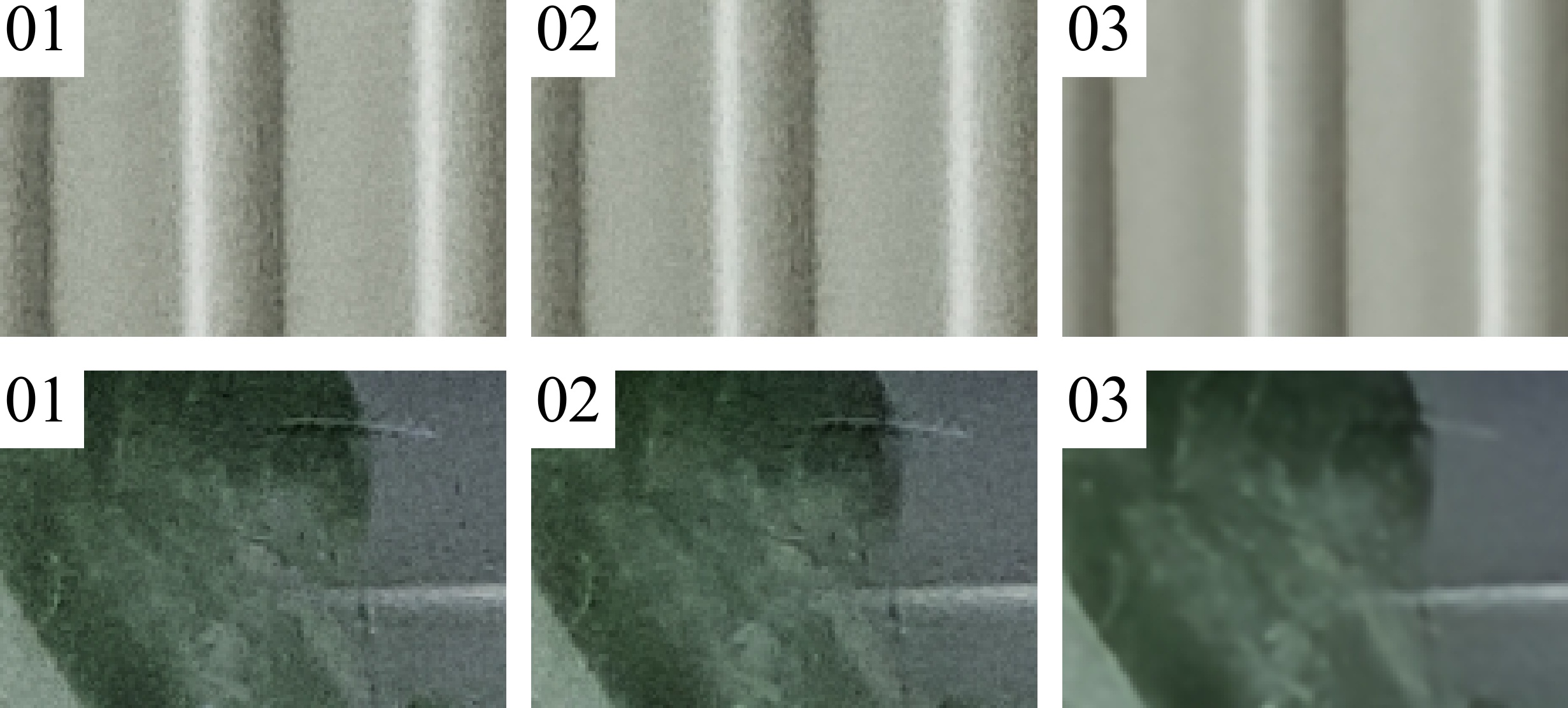}
  \end{subfigure}%
  \vspace{0.1cm}
  \begin{subfigure}[t]{0.48\textwidth}
    \centering
    \includegraphics[width=\textwidth]{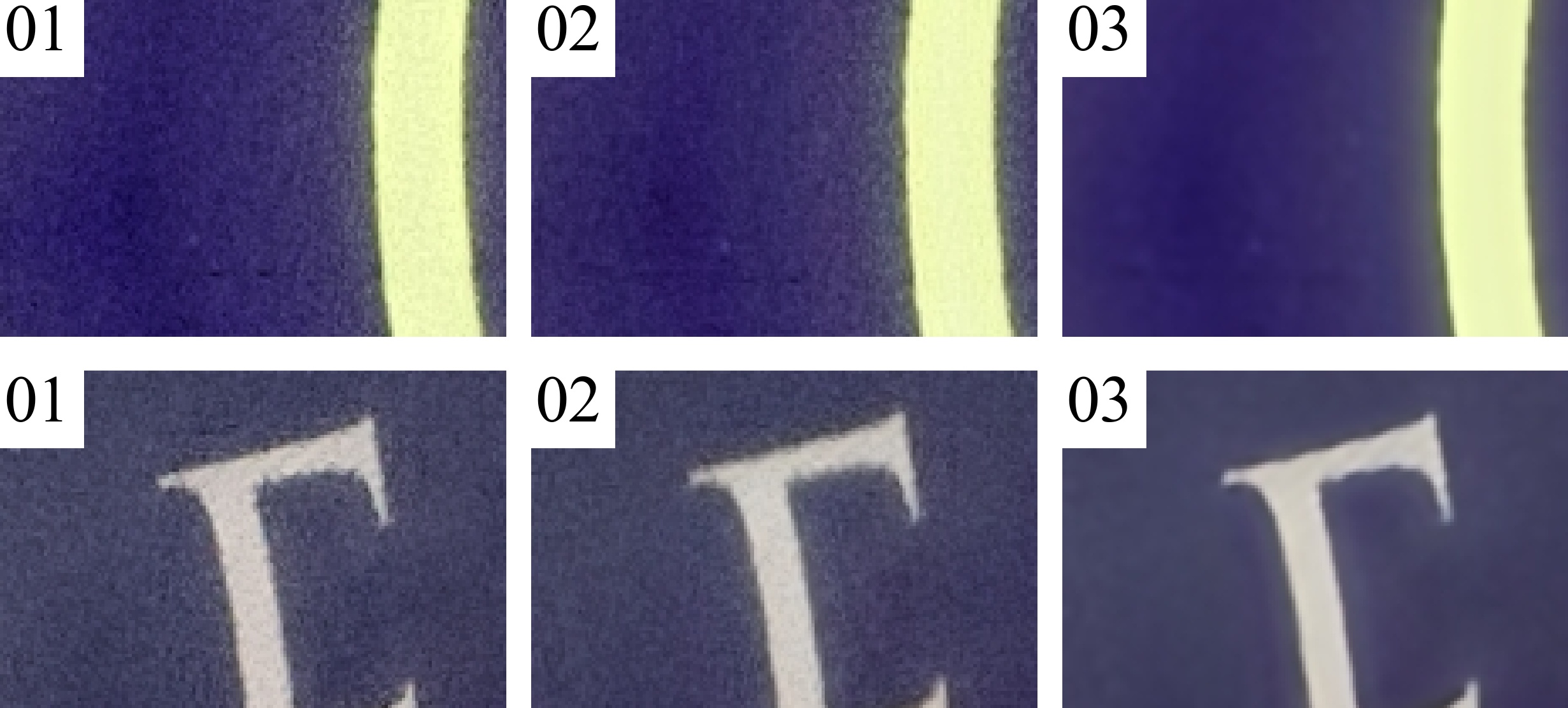}
  \end{subfigure}%
  % \vspace{0.1cm}
  % \begin{subfigure}[t]{0.46\textwidth}
  %   \centering
  %   \includegraphics[width=\textwidth]{figures/results_data_augmentation_chairs2.jpg}
  % \end{subfigure}%
  \caption{Result of HeliconFocus, FocusDeep architecture, and FocusDeep architecture with noise reduction on our raw dataset.}
  \label{figure:noise}
  \vspace{-0.4cm}
\end{figure}

\paragraph{Denoising with Data Augmentation.}

Creating synthetic focus bracketing bursts from all-in-focus real images is extremely challenging, and we have chosen instead to use HeliconFocus as (pseudo) ground truth and train our algorithm on real bursts acquired with focus bracketing.
While we hypothesize that a traditional training approach would be able to outperform HeliconFocus, we demonstrate that it is possible to achieve superior qualitative results by adapting the training algorithm with regularization or data augmentation.
In this regard, we conducted experiments by integrating realistic synthetic noise as a data augmentation technique during training.

Figures~\ref{figure:results_deep_learning_test}~and~\ref{figure:results_deep_learning_train} are proofs of concept that we can achieve higher qualitative results with a deep learning architecture, demonstrating improvement over HeliconFocus.
To qualitatively improve upon HeliconFocus, we train a FocusDeep model to jointly demosaic, focus stack and denoise raw bursts.
In order to denoise directly from raw busts, we rely on the approach proposed by~\cite{brooks2019unprocessing} to synthesize realistic raw sensor measurements.
The noisy raw values can be considered as a heteroscedastic Gaussian and defined each pixel intensity $y$ as 
$y \sim \mathcal N(\mu=x, \sigma^2=\lambda_\text{read} + \lambda_\text{shot} x)$
where $x$ is the original pixel value, $\lambda_\text{shot}$ is defined as a Poisson random variable whose mean is the true light intensity, and $\lambda_\text{read}$ is defined as a Gaussian random variable with zero mean and fixed variance.
Instead of training on synthetic raw, as it was done by~\cite{brooks2019unprocessing}, we directly add a reasonable level of noise in our real raw bursts during training following the same sampling procedure of shot/read noise factors: 
{\small
\begin{align*}
  &\log(\lambda_\text{shot}) \sim {\mathcal{U}}(a=\log(0.0001),b=\log(0.012)) \\
  &\log(\lambda_\text{read}) \mid \log(\lambda_\text{shot}) \sim \\
  & \quad\quad\quad\quad {\mathcal{N}}(\mu=2.18 \log(\lambda_\text{shot})+ 1.2 ,\sigma=0.26) .
\end{align*}
}
Figure~\ref{figure:noise} presents the result of this approach with real noise from the raw burst.
It shows the result of HeliconFocus, our FocusDeep architecture and our FocusDeep architecture with added noise in that order respectively. 
We observe a high reduction of noise between the last column compared to the first two.
Figure~\ref{fig:teaser} presents similar results with 30 focus-bracketed raw images taken by an iPhone 12 at 1000 ISO. In this short-exposure, high-ISO setting aimed at minimizing motion blur, our solution recovers the same level of detail as HeliconFocus, but achieves a significant level of noise reduction.

\begin{figure}[t]
  \centering
  \includegraphics[width=0.48\textwidth,trim={0 0 0 10cm},clip]{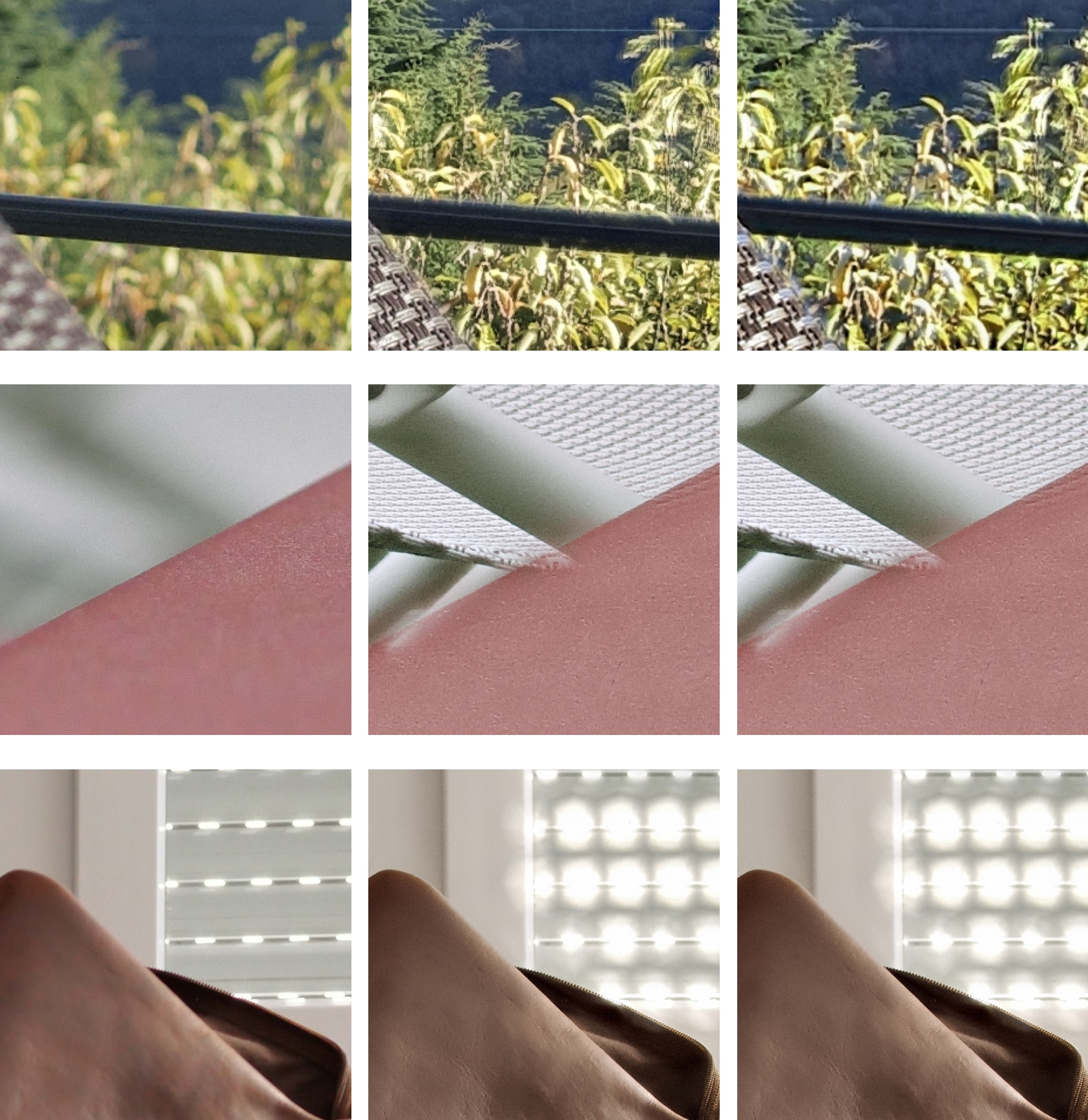}
  \caption{This figure shows some failure modes of HeliconFocus, although HeliconFocus is a state-of-the-art software for focus stacking, this figure demonstrates the difficulty of obtaining a perfect focus stacking in difficult conditions. Left column: real images, middle column: result of HeliconFocus and right column: result of FocusDeep. At the top, we can observe a superposition between the foreground and background not respecting the real depth of the scene. At the bottom, we can observe that HeliconFocus confuses blurry pixels with sharp ones in a saturated setting.}
  \label{figure:fail_helicon}
  \vspace{-0.2cm}
\end{figure}

\section{Future Work}

\paragraph{Architecture and Training.}

In the context of micro and macro photography, the depth of field is so shallow that to perform a successful focus stacking, hundreds of images are required.
We observe on the images taken with the Olympus lens (Figure~\ref{figure:burst-dataset} in the Appendix) that some areas of the image are still blurry despite performing focus stacking with 30 images.
In this context, our approach based on supervised learning could exhibit some limitations.
Indeed, while inference can always be computed with a window sliding algorithm, training on extended bursts may prove difficult due to the limits of GPUs memory.
Furthermore, our FocusDeep architecture is devised with a fixed burst length and is not invariant to the order of the image in the burst. 
Indeed, the order of the images of the burst offers real information about the depth of the scene, it could be interesting to investigate deep learning architectures for focus stacking with an arbitrary number of inputs and order invariance, as in~\cite{xiao2022general}.

Currently, our approach requires that all images of the burst be aligned.
Although our FocusDeep architecture has a feature alignment block, it is not robust to large misalignment due to the magnification of focus-bracketed raw images.
An end-to-end approach should be investigated, it would require the network to have a ``global view'' of the images of the burst instead of a local one as with our $128 \times 128$ crops.

\paragraph{Pseudo Ground Truth.}

In this work, we chose to design a dataset with real training data and pseudo ground truth, whereas previous work chose to devise synthetic training data based on a real all-in-focus image.
Although we have shown that the quality of this approach could prove useful to photographers, it cannot be perfect due to the errors made by the software used to generate the ground truth.
Indeed, since the geometry of a camera changes during focus bracketing, it may very well be that the same image pixel is the projection of multiple in-focus scene points, along different light rays of course for scenes made of opaque solids. 
This illustrates the intrinsic difficulty of focus stacking, where finding correspondences between patches with vastly different levels of blur and aligning them is extremely difficult, with the added difficulty that a pixel may be the projection of several in-focus scene points, as illustrated by Figure~\ref{figure:fail_helicon}.
In the context of supervised learning, our deep learning architecture will learn to reproduce the same mistakes.
Further investigation with different software for generating the ground truth could be investigated.
Also, a promising approach would be to use unsupervised learning algorithms which do not require any ground truth, however, devising a proper loss function might be challenging.

\section{Conclusion}

This paper introduces a large-scale dataset with bursts from focus-bracketed raw images for focus stacking. We also introduce a FocusDeep architecture that, when trained on our new dataset, can perform joint demosaic and focus stacking tasks as well as the best commercial software available. 
Furthermore, by adding realistic noise during training, we can show that our approach is more robust to noise than the method used to generate the ground truth.
We believe that this dataset combined with supervised learning approaches can significantly outperform existing focus stacking solutions.
This work goes beyond focus stacking as we strongly believe that our dataset can be used in many other tasks.
For example, our dataset could be used in the context of depth from defocus~\cite{subbarao1994depth,hazirbas2018deep}.
Indeed, the blur in each image gives information about the depth of the scene.
Furthermore, our dataset could be used to learn how to perform realistic simulated blur and simulated depth-of-field, which is a highly investigated topic~\cite{wadhwa2018synthetic}.

\newpage
\subsubsection*{Acknowledgment}

This work was granted access to the HPC resources of IDRIS under the allocation AD011013259 made by GENCI.

%%%%%%%%% REFERENCES
{\small
\bibliographystyle{ieee_fullname}
\bibliography{bibliography}
}
\clearpage
\appendix

\twocolumn[

\section{Appendix}

\paragraph{Additional Training details.}

We train the UNet and FocusDeep architectures on our RGB and raw datasets with RGB pseudo ground truth.
Our inputs are either 4-channels `RGGB' raw bursts for the raw dataset and 3-channels RGB bursts for the RGB dataset, the outputs are RGB in both cases.
We use a $\ell_1$ loss between the output and the ground truth and we ignore 4 pixels around the border.
We use Adam optimizer with an initial learning rate set to $10^{-4}$, reduced by $50\%$ every 100 epochs.
We train our networks for 300 epochs, which takes around 3 days on 16 GPUs V100.

]

\begin{figure*}[b]
  \centering
  \begin{subfigure}[t]{1\textwidth}
    \centering
    \includegraphics[width=0.95\textwidth]{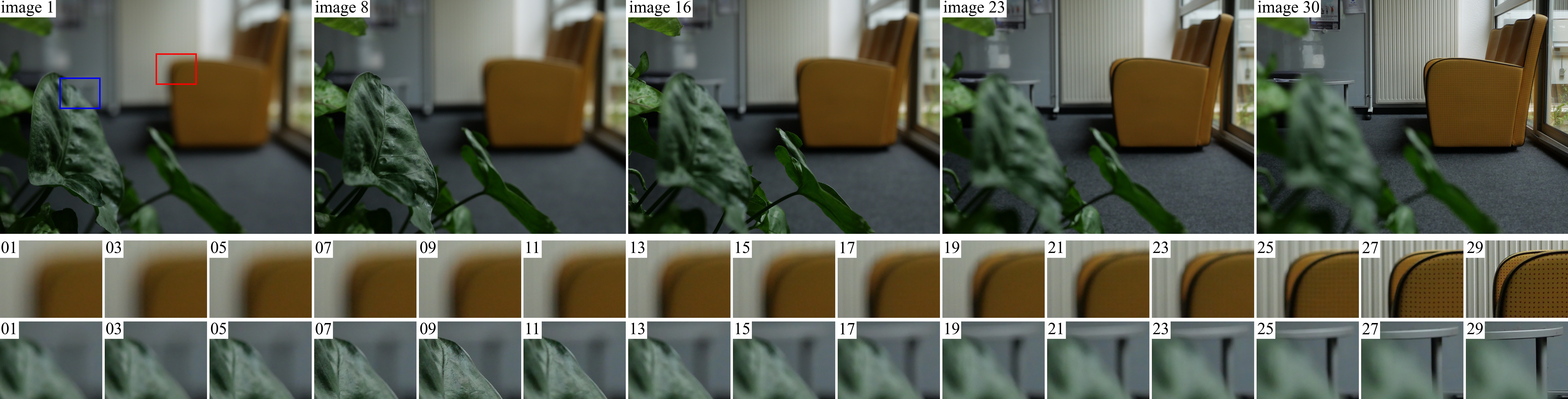}
  \end{subfigure}%
  \vspace{0.05cm}
  \begin{subfigure}[t]{1\textwidth}
    \centering
    \includegraphics[width=0.95\textwidth]{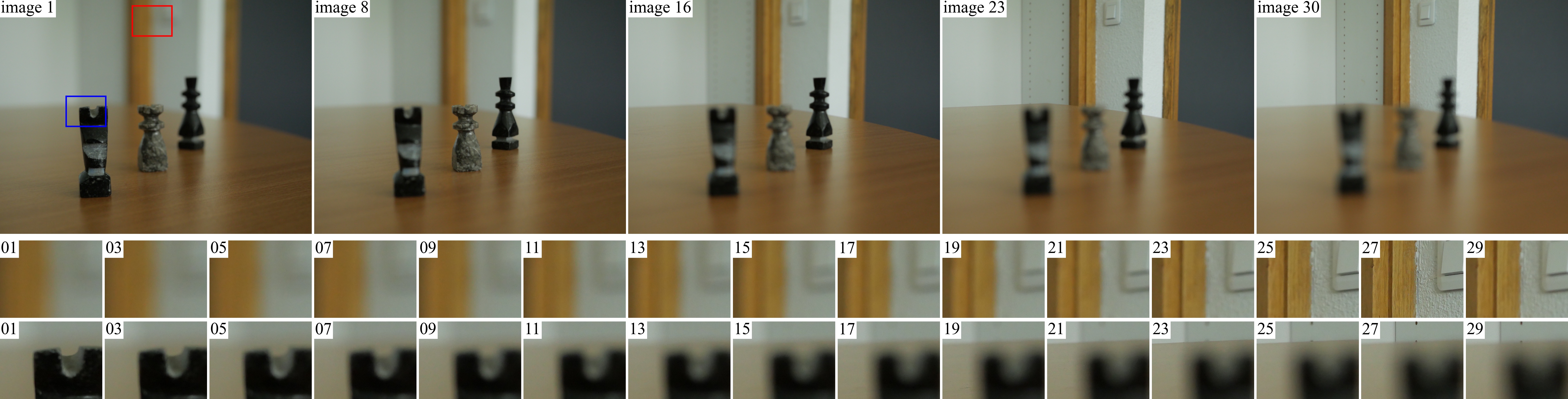}
  \end{subfigure}%
  \vspace{0.05cm}
  \begin{subfigure}[t]{1\textwidth}
    \centering
    \includegraphics[width=0.95\textwidth]{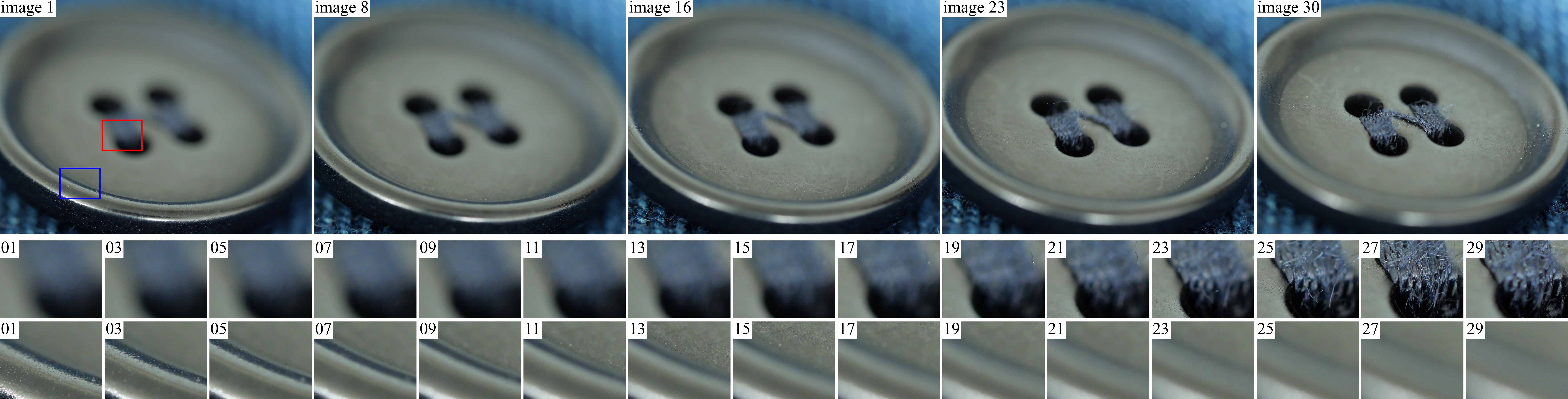}
  \end{subfigure}%
  \vspace{0.05cm}
  \begin{subfigure}[t]{1\textwidth}
    \centering
    \includegraphics[width=0.95\textwidth]{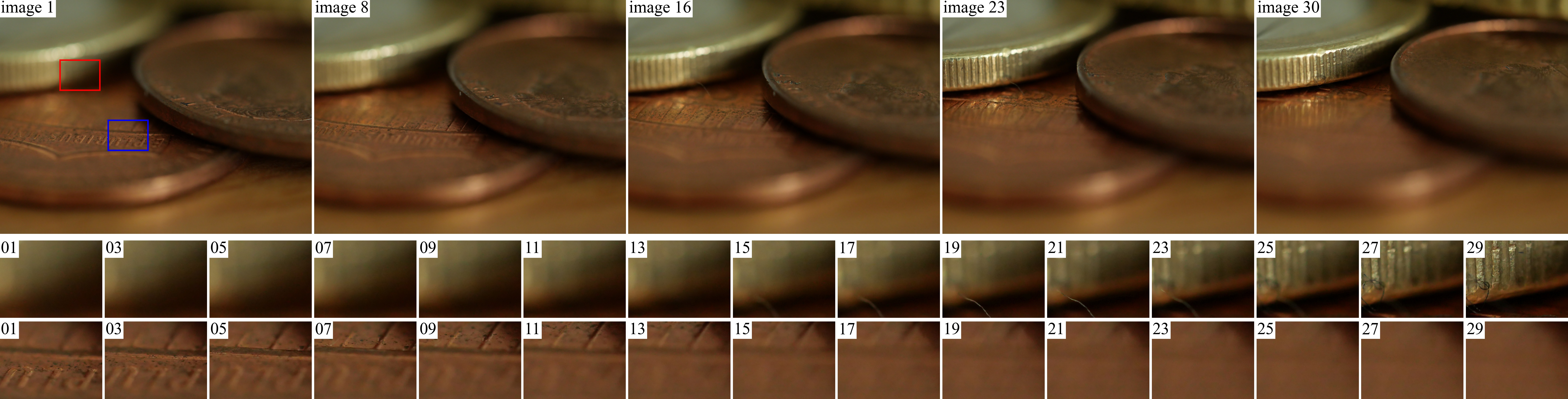}
  \end{subfigure}
  \caption{This figure shows different bursts from our train dataset taken with a Panasonic Lumix DC-GX9. The number on the images corresponds to the position of the image in the burst. The first two bursts were taken with the Leica 25mm F1.4 II lens. They both represent a scene in a room with objects in the foreground and background. The last two bursts were taken with the Olympus M.60mm F2.8 Macro lens and represent very small objects.}
  \label{figure:burst-dataset}
\end{figure*}

\begin{figure*}[t]
  \centering
  \begin{subfigure}[t]{1\textwidth}
    \centering
    \includegraphics[width=\textwidth]{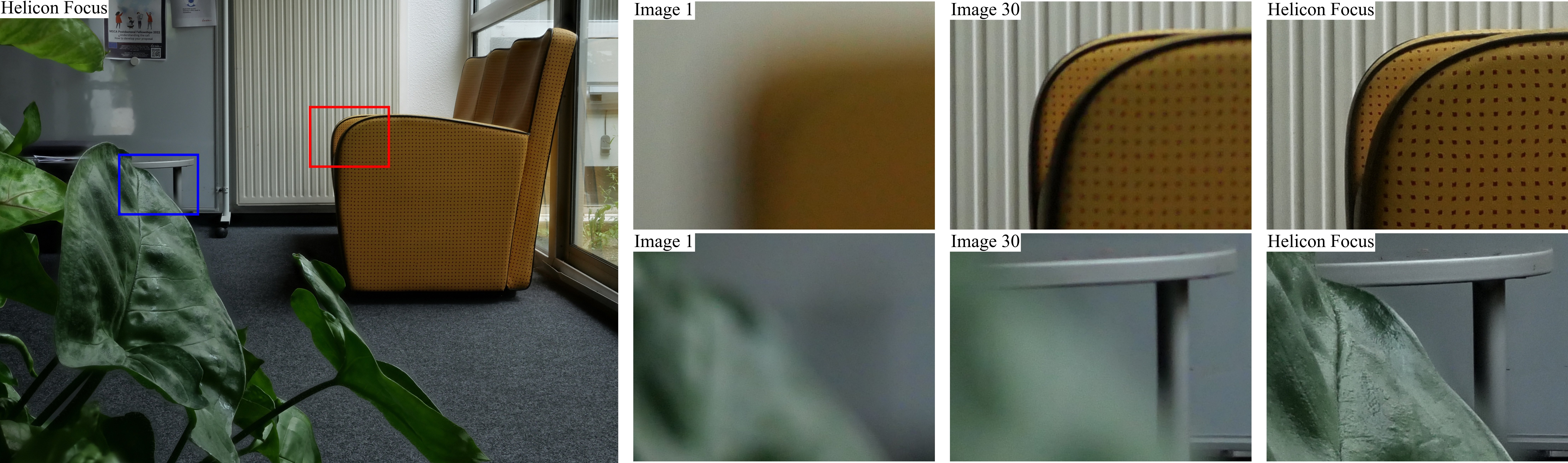}
  \end{subfigure}%
  \vspace{0.1cm}
  \begin{subfigure}[t]{1\textwidth}
    \centering
    \includegraphics[width=\textwidth]{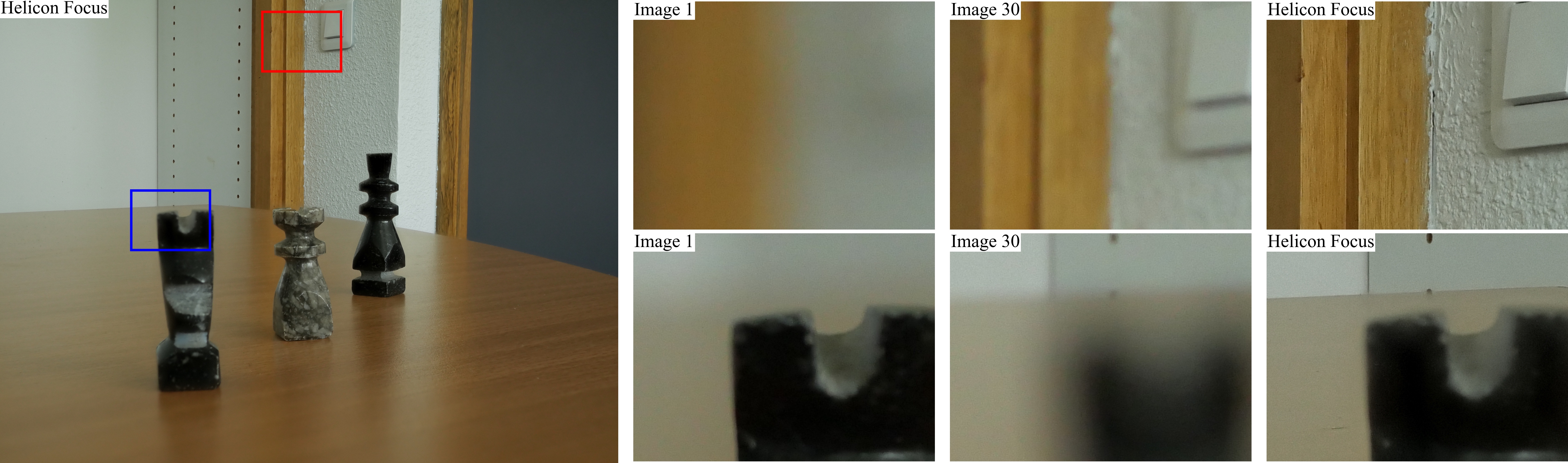}
  \end{subfigure}%
  \vspace{0.1cm}
  \begin{subfigure}[t]{1\textwidth}
    \centering
    \includegraphics[width=\textwidth]{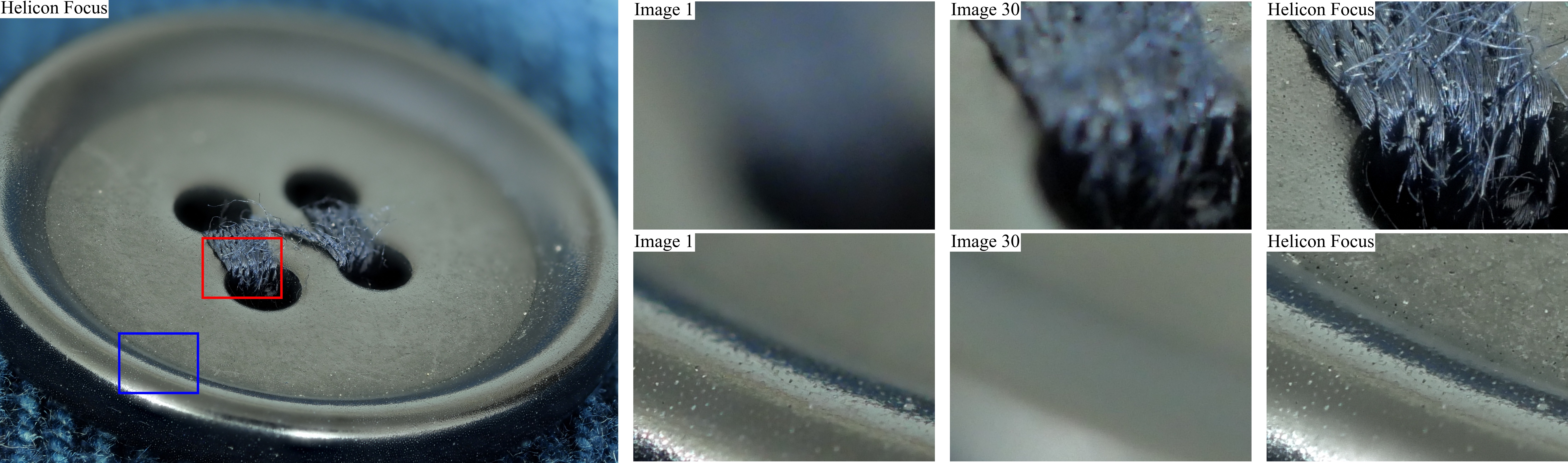}
  \end{subfigure}%
  \vspace{0.1cm}
  \begin{subfigure}[t]{1\textwidth}
    \centering
    \includegraphics[width=\textwidth]{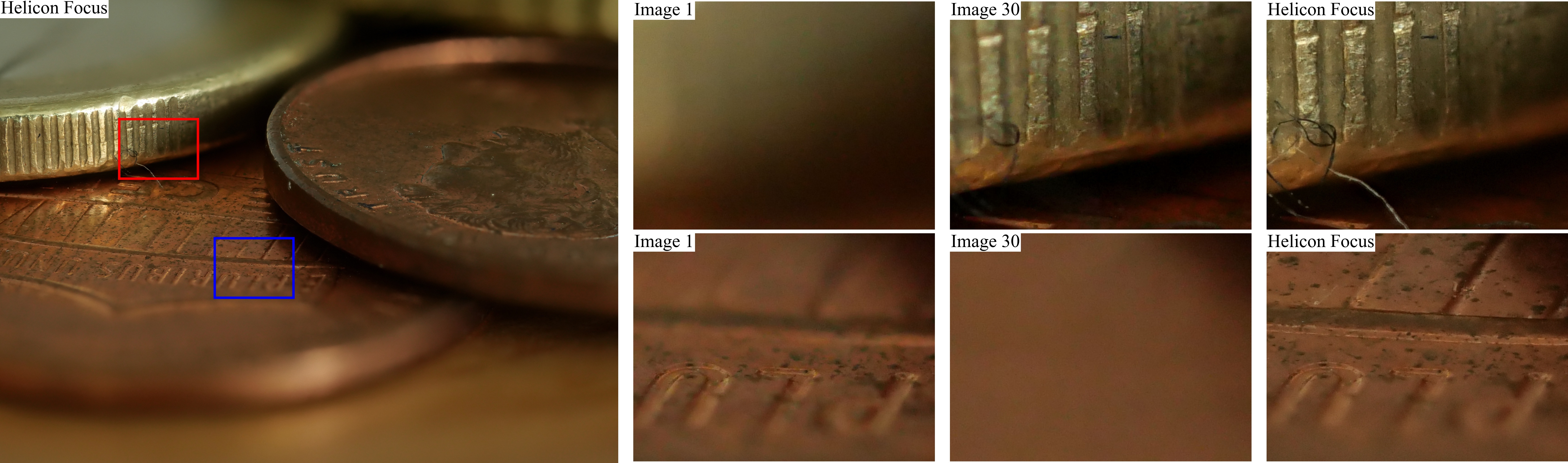}
  \end{subfigure}
  \caption{This figure shows the result of HeliconFocus on the four bursts presented in Figure~\ref{figure:burst-dataset}. The crops show a zoom on the first and last image of the burst as well as the same zoom on the Helicon Focus result.}
  \label{figure:burst-heliconfocus}
\end{figure*}

\begin{figure*}[t]
  \centering
  \begin{subfigure}[t]{1\textwidth}
    \centering
    \includegraphics[width=0.99\textwidth]{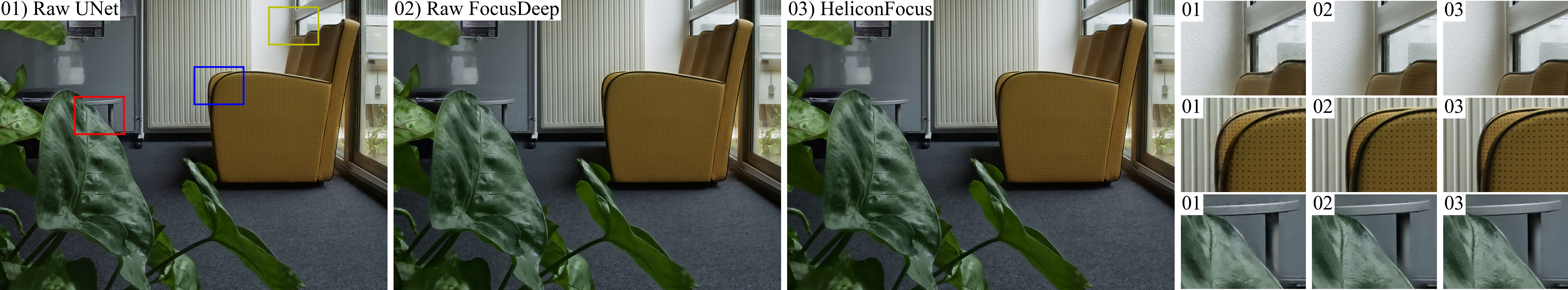}
  \end{subfigure}%
  \vspace{0.1cm}
  \begin{subfigure}[t]{1\textwidth}
    \centering
    \includegraphics[width=0.99\textwidth]{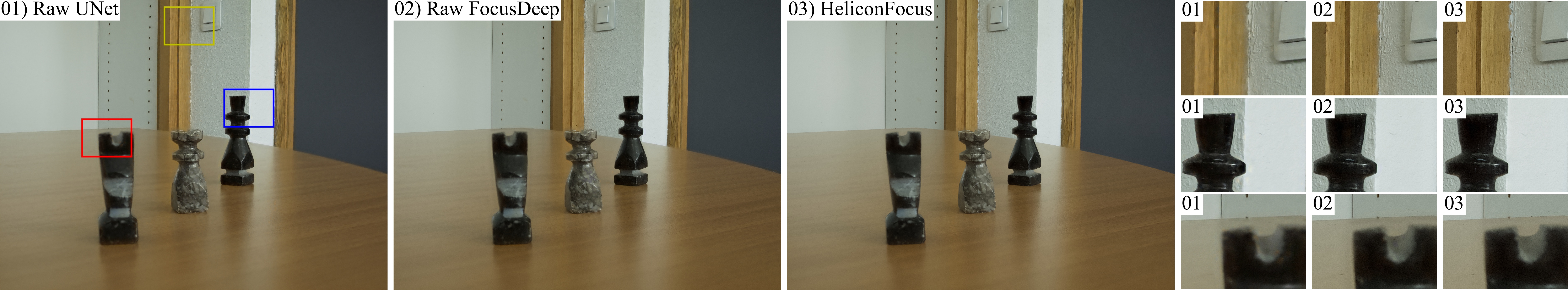}
  \end{subfigure}%
  \vspace{0.1cm}
  \begin{subfigure}[t]{1\textwidth}
    \centering
    \includegraphics[width=0.99\textwidth]{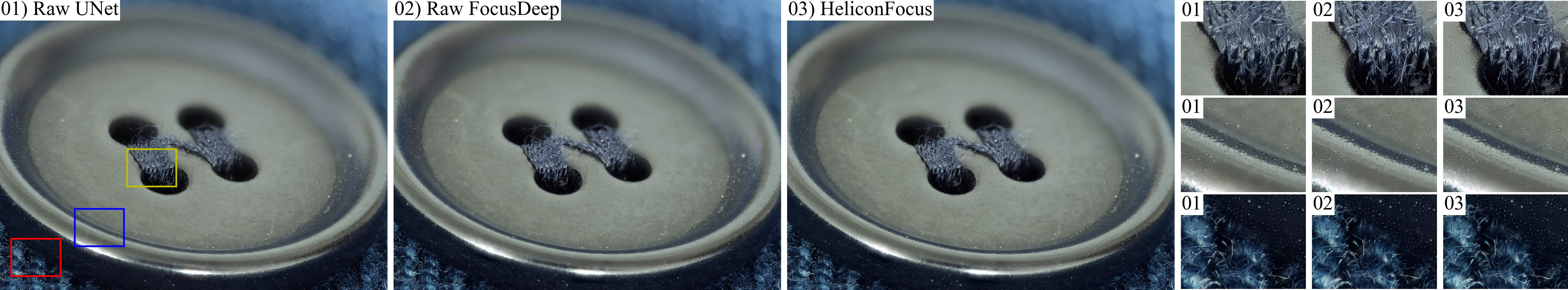}
  \end{subfigure}%
  \vspace{0.1cm}
  \begin{subfigure}[t]{1\textwidth}
    \centering
    \includegraphics[width=0.99\textwidth]{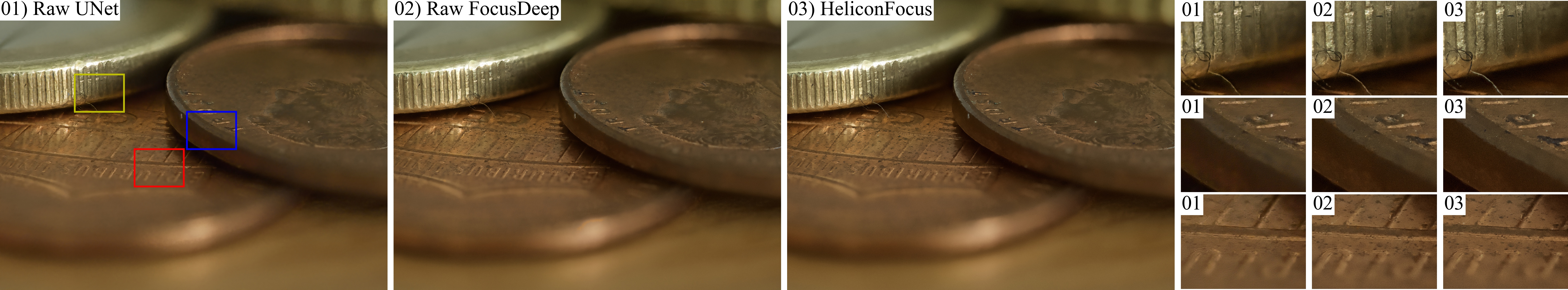}
  \end{subfigure}
  \caption{This figure presents results on training data of the UNet and FocusDeep architecture with raw bursts. We observe some small artefacts on the results of the UNet architecture confirming the lower quantitative results.}
  \label{figure:results_deep_learning_train}
\end{figure*}

\end{document}

%% file: definitions.tex
\usepackage{mathtools}      % Mathematical tools to use with amsmath
\usepackage{amssymb}
\usepackage{mathdots}       % dots commands for matrices 
\usepackage{amsfonts}       % blackboard math symbols
\usepackage{nicefrac}       % compact symbols for 1/2, etc.
\usepackage{physics}        % typesetting equations for physics simpler
\usepackage{bm}             % Access bold symbols in maths mode
\usepackage{centernot}      % The package provides \centernot
\usepackage{thmtools}       % Extensions to theorem environments
\usepackage{amsmath}

\def\ddefloop#1{\ifx\ddefloop#1\else\ddef{#1}\expandafter\ddefloop\fi}

\def\ddef#1{\expandafter\def\csname #1bb\endcsname{\ensuremath{\mathbb{#1}}}}
\ddefloop ABCDEFGHIJKLMNOPQRSTUVWXYZ\ddefloop

\def\ddef#1{\expandafter\def\csname #1set\endcsname{\ensuremath{\mathcal{#1}}}}
\ddefloop ABCDEFGHIJKLMNOPQRSTUVWXYZ\ddefloop

\def\ddef#1{\expandafter\def\csname #1mat\endcsname{\ensuremath{\mathbf{#1}}}}
\ddefloop ABCDEFGHIJKLMNOPQRSTUVWXYZ\ddefloop

\def\ddef#1{\expandafter\def\csname #1matsf\endcsname{\ensuremath{\mathsf{#1}}}}
\ddefloop ABCDEFGHIJKLMNOPQRSTUVWXYZ\ddefloop

\def\ddef#1{\expandafter\def\csname #1vec\endcsname{\ensuremath{\mathbf{#1}}}}
\ddefloop abcdefghijklmnopqrstuvwxyz\ddefloop

%% file: figures/datasets.tex
\tikzset{%
  >={Latex[width=0mm,length=0mm]},
  base/.style = {fill=white, font=\sffamily},
  bayer/.style={
    draw=black,
    rounded corners=0,
    line width=0.5pt,
    minimum width=0.15cm,
    minimum height=0.11cm
  },
  burst/.style={
    draw=black,
    rounded corners=0,
    line width=1pt,
    minimum width=1.5cm,
    minimum height=1.12cm
  },
  crops/.style={
    draw=black,
    rounded corners=0,
    line width=1pt,
    minimum width=0.75cm,
    minimum height=0.56cm
  },
  rectangle1/.style={
    rounded corners,
    fill=myorange,
    draw=black,
    line width=1pt,
    minimum width=2cm,
    minimum height=1cm
  },
  descr/.style={
    fill=white,
    inner sep=2.5pt
  },
  connector/.style={
    -latex,
    color=black,
    line width=1pt,
  },
}

\def\offset{0.22}

\begin{tikzpicture}[every node/.style=base, align=center, scale=5]

  % RGGB raw burst
  \node[rectangle1, fill=none, draw=none] (raw) at (-0.8, -0.7) {%
    \begin{tikzpicture}
       \foreach \i in {1,3,5}
       {
	 \node[bayer, fill=burstgreen] at (-0.7+0*\offset, -0.3+\i*\offset) {};
	 \node[bayer, fill=burstblue]  at (-0.7+1*\offset, -0.3+\i*\offset) {};
	 \node[bayer, fill=burstgreen] at (-0.7+2*\offset, -0.3+\i*\offset) {};
	 \node[bayer, fill=burstblue]  at (-0.7+3*\offset, -0.3+\i*\offset) {};
	 \node[bayer, fill=burstgreen] at (-0.7+4*\offset, -0.3+\i*\offset) {};
	 \node[bayer, fill=burstblue]  at (-0.7+5*\offset, -0.3+\i*\offset) {};
	 \node[bayer, fill=burstgreen] at (-0.7+6*\offset, -0.3+\i*\offset) {};
	 \node[bayer, fill=burstblue]  at (-0.7+7*\offset, -0.3+\i*\offset) {};
       }
       \foreach \i in {0,2,4,6}
       {
	 \node[bayer, fill=burstred]   at (-0.7+0*\offset, -0.3+\i*\offset) {};
	 \node[bayer, fill=burstgreen] at (-0.7+1*\offset, -0.3+\i*\offset) {};
	 \node[bayer, fill=burstred]   at (-0.7+2*\offset, -0.3+\i*\offset) {};
	 \node[bayer, fill=burstgreen] at (-0.7+3*\offset, -0.3+\i*\offset) {};
	 \node[bayer, fill=burstred]   at (-0.7+4*\offset, -0.3+\i*\offset) {};
	 \node[bayer, fill=burstgreen] at (-0.7+5*\offset, -0.3+\i*\offset) {};
	 \node[bayer, fill=burstred]   at (-0.7+6*\offset, -0.3+\i*\offset) {};
	 \node[bayer, fill=burstgreen] at (-0.7+7*\offset, -0.3+\i*\offset) {};
       }

       \foreach \i in {1,3,5}
       {
	 \node[bayer, fill=burstgreen] at (-3.2+0*\offset, -0.3+\i*\offset) {};
	 \node[bayer, fill=burstblue]  at (-3.2+1*\offset, -0.3+\i*\offset) {};
	 \node[bayer, fill=burstgreen] at (-3.2+2*\offset, -0.3+\i*\offset) {};
	 \node[bayer, fill=burstblue]  at (-3.2+3*\offset, -0.3+\i*\offset) {};
	 \node[bayer, fill=burstgreen] at (-3.2+4*\offset, -0.3+\i*\offset) {};
	 \node[bayer, fill=burstblue]  at (-3.2+5*\offset, -0.3+\i*\offset) {};
	 \node[bayer, fill=burstgreen] at (-3.2+6*\offset, -0.3+\i*\offset) {};
	 \node[bayer, fill=burstblue]  at (-3.2+7*\offset, -0.3+\i*\offset) {};
       }
       \foreach \i in {0,2,4,6}
       {
	 \node[bayer, fill=burstred]   at (-3.2+0*\offset, -0.3+\i*\offset) {};
	 \node[bayer, fill=burstgreen] at (-3.2+1*\offset, -0.3+\i*\offset) {};
	 \node[bayer, fill=burstred]   at (-3.2+2*\offset, -0.3+\i*\offset) {};
	 \node[bayer, fill=burstgreen] at (-3.2+3*\offset, -0.3+\i*\offset) {};
	 \node[bayer, fill=burstred]   at (-3.2+4*\offset, -0.3+\i*\offset) {};
	 \node[bayer, fill=burstgreen] at (-3.2+5*\offset, -0.3+\i*\offset) {};
	 \node[bayer, fill=burstred]   at (-3.2+6*\offset, -0.3+\i*\offset) {};
	 \node[bayer, fill=burstgreen] at (-3.2+7*\offset, -0.3+\i*\offset) {};
       }
       \path[draw=black, line width=1pt, dashed] (-3.6, -0.7) rectangle (1.2, 1.4);
       \node[fill=none] at (-1.15, +0.3) {$\boldsymbol{\cdots}$};
       \node[fill=none] at (-1.20, -1.3) {{\small Raw Burst} \\ {\small 30 x 5184 x 3888 x 1}};
    \end{tikzpicture}
  };

  % RGGB raw burst
  \node[rectangle1, fill=none, draw=none] (rggb) at (0.8, -0.7) {%
    \begin{tikzpicture}
       \node[burst, fill=burstred]   at (-0.6, +0.6) {};
       \node[burst, fill=burstgreen] at (-0.4, +0.4) {};
       \node[burst, fill=burstgreen] at (-0.2, +0.2) {};
       \node[burst, fill=burstblue]  at (-0.0, +0.0) {};
       \node[burst, fill=burstred]   at (-3.6, +0.6) {};
       \node[burst, fill=burstgreen] at (-3.4, +0.4) {};
       \node[burst, fill=burstgreen] at (-3.2, +0.2) {};
       \node[burst, fill=burstblue]  at (-3.0, +0.0) {};
       \path[draw=black, line width=1pt, dashed] (-4.8, -0.9) rectangle (1.2, 1.5);
       \node[fill=none] at (-1.8, +0.0) {$\boldsymbol{\cdots}$};
       \node[fill=none] at (-1.6, -1.4) {{\small RGGB Burst} \\ {\small 30 x 2592 x 1944 x 4}};
    \end{tikzpicture}
  };

  % RGB burst
  \node[rectangle1, fill=none, draw=none] (rgb) at (0.8, 0.0) {%
    \begin{tikzpicture}
       \node[burst, fill=burstred]   at (-0.4, +0.4) {};
       \node[burst, fill=burstgreen] at (-0.2, +0.2) {};
       \node[burst, fill=burstblue]  at (-0.0, +0.0) {};
       \node[burst, fill=burstred]   at (-3.4, +0.4) {};
       \node[burst, fill=burstgreen] at (-3.2, +0.2) {};
       \node[burst, fill=burstblue]  at (-3.0, +0.0) {};
       \path[draw=black, line width=1pt, dashed] (-4.8, -0.9) rectangle (1.0, 1.2);
       \node[fill=none] at (-1.7, +0.0) {$\boldsymbol{\cdots}$};
       \node[fill=none] at (-1.6, -1.4) {{\small RGB Burst} \\ {\small 30 x 5184 x 3888 x 3}};
    \end{tikzpicture}
  };

  % RGB crops with RGB ground-truth
  \node[rectangle1, fill=none, draw=none] (crops rgb) at (3.0, +0.6) {%
    \begin{tikzpicture}
       \path[draw=black, line width=1pt, dashed] (-3.2, -0.7) rectangle (0.8, 1.2);
       \node[crops, fill=burstred]   at (-0.4, +0.4) {};
       \node[crops, fill=burstgreen] at (-0.2, +0.2) {};
       \node[crops, fill=burstblue]  at (-0.0, +0.0) {};
       \node[crops, fill=burstred]   at (-2.4, +0.4) {};
       \node[crops, fill=burstgreen] at (-2.2, +0.2) {};
       \node[crops, fill=burstblue]  at (-2.0, +0.0) {};
       \node[fill=none] at (-1.2, +0.0) {$\boldsymbol{\cdots}$};
       \node[fill=none] at (-1.2, -1.42) {%
         {\small RGB crops with RGB ground-truth} \\
         {\footnotesize 1200 x 30 x 128 x 128 x 3} \\
         {\footnotesize 1200 x 1 x 128 x 128 x 3}
       };
    \end{tikzpicture}
  };

  \node[rectangle1, fill=none, draw=none] (crops raw) at (3.0, -0.7) {%
    \begin{tikzpicture}
       \path[draw=black, line width=1pt, dashed] (-3.2, -0.7) rectangle (0.8, 1.2);
       \node[crops, fill=burstred]   at (-0.6, +0.6) {};
       \node[crops, fill=burstgreen] at (-0.4, +0.4) {};
       \node[crops, fill=burstgreen] at (-0.2, +0.2) {};
       \node[crops, fill=burstblue]  at (-0.0, +0.0) {};
       \node[crops, fill=burstred]   at (-2.6, +0.6) {};
       \node[crops, fill=burstgreen] at (-2.4, +0.4) {};
       \node[crops, fill=burstgreen] at (-2.2, +0.2) {};
       \node[crops, fill=burstblue]  at (-2.0, +0.0) {};
       \node[fill=none] at (-1.2, +0.0) {$\boldsymbol{\cdots}$};
       \node[fill=none] at (-1.2, -1.5) {%
         {\small RGGB crops with RGB ground-truth} \\
         {\footnotesize 1200 x 30 x 64 x 64 x 4} \\
         {\footnotesize 1200 x 1 x 128 x 128 x 3}
       };
    \end{tikzpicture}
  };

  \draw[decoration={brace,mirror},decorate,line width=1pt] (2.65,-0.87) -- (2.65, -1.03);
  \draw[decoration={brace,mirror},decorate,line width=1pt] (2.65,+0.43) -- (2.65, +0.27);

  \node[rectangle1] (demosaicing) at (-0.3, +0.0) {Demosaicing};
  \node[rectangle1, minimum width=3.5cm] (rgb registration) at (2.0, +0.0) {RGB registration};
  \node[rectangle1, minimum width=3.5cm] (raw registration) at (2.0, -0.7) {Raw registration from \\ affine transforms};
  \node[rectangle1] (helicon focus) at (3.0, +0.0) {Fusion with \\ \textbf{HeliconFocus}};
  \node[rectangle1, fill=none, draw=none] (helicon sub) at (3.0, -0.15) {{\small 1 x 5184 x 3888 x 3}};

  \draw[connector] (raw) |- (demosaicing);
  \draw[connector] (demosaicing) -- (rgb);
  \draw[connector] (rgb) -- (rgb registration);
  \draw[connector] (rgb registration) -- (helicon focus);
  \draw[connector, -latex] (rgb registration) -- node[descr, pos=0.5] {Transforms \\ computed from RGB} (raw registration);
  \draw[connector] (raw) -- (rggb);
  \draw[connector] (rggb) -- (raw registration);

  \draw[connector] (rgb registration) |- (crops rgb);
  \draw[connector] (raw registration) -- (crops raw);

  \draw[connector] (helicon focus) -- (crops rgb);
  \draw[connector] (helicon sub) -- (crops raw);

\end{tikzpicture}%

%% file: figures/architecture.tex
\tikzset{%
  >={Latex[width=0mm,length=0mm]},
  base/.style = {fill=white, font=\sffamily},
  burst/.style={
    draw=black,
    rounded corners=0,
    line width=1pt,
    minimum width=1.5cm,
    minimum height=1.12cm
  },
  crops/.style={
    draw=black,
    rounded corners=0,
    line width=1pt,
    minimum width=0.75cm,
    minimum height=0.56cm
  },
  rectangle1/.style={
    rounded corners,
    fill=myorange,
    draw=black,
    line width=1pt,
    minimum width=2cm,
    minimum height=1cm
  },
  descr/.style={
    fill=white,
    inner sep=2.5pt
  },
  connector/.style={
    -latex,
    color=black,
    line width=1pt,
  },
  rectangle connector/.style={
    connector,
    to path={(\tikztostart) |- ++(3.1,0.25) \tikztonodes -- (\tikztotarget)},
    pos=0.5
  },
  % rectangle connector/.default=-2cm,
}
\begin{tikzpicture}[every node/.style=base, align=center, scale=5]

  \node[rectangle1, fill=none, draw=none] (crops raw) at (0.0, -0.1) {%
    \begin{tikzpicture}
       \path[draw=black, line width=1pt, dashed] (-3.2, -0.7) rectangle (0.8, 1.2);
       \node[crops, fill=burstred]   at (-0.6, +0.6) {};
       \node[crops, fill=burstgreen] at (-0.4, +0.4) {};
       \node[crops, fill=burstgreen] at (-0.2, +0.2) {};
       \node[crops, fill=burstblue]  at (-0.0, +0.0) {};
       \node[crops, fill=burstred]   at (-2.6, +0.6) {};
       \node[crops, fill=burstgreen] at (-2.4, +0.4) {};
       \node[crops, fill=burstgreen] at (-2.2, +0.2) {};
       \node[crops, fill=burstblue]  at (-2.0, +0.0) {};
       \node[fill=none] at (-1.2, +0.0) {$\boldsymbol{\cdots}$};
       \node[fill=none] at (-1.2, -1.2) {128x128 Raw crops};
    \end{tikzpicture}
  };

  \node[rectangle1, fill=none, draw=none] (final) at (3.8, -0.1) {%
    \begin{tikzpicture}
       \pgftext{\includegraphics[scale=0.2]{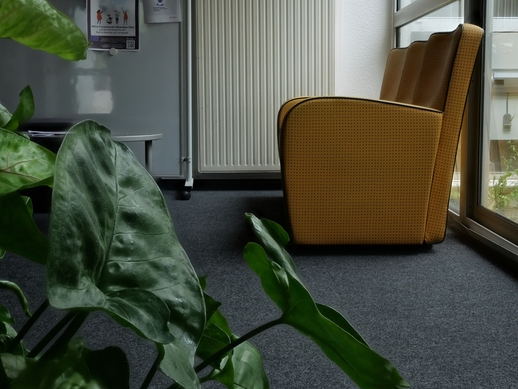}} at (0, 0);
       \node[fill=none] at (0.0, -2.0) {All-in-focus image (RGB)};
    \end{tikzpicture}
  };

  \node[rectangle1] (Feature Extractor) at (0.8, 0.0) {Feature \\ Extractor};
  \node[rectangle1] (FEPCD Align) at (1.4, 0.0) {Alignement};
  \node[rectangle1] (Fusion) at (2.0, 0.0) {Conv};
  \node[rectangle1] (LRCN) at (2.6, 0.0) {LRCN};
  \node[circle, draw] (circle) at (3.1, 0.0) {$+$};

  \node[rectangle] at (1.10, -0.3) {(a) Feature Alignment};
  \node[rectangle] at (2.00, -0.3) {(b) Fusion};
  \node[rectangle] at (2.75, -0.3) {(c) Reconstruction};

  \draw[connector] ([yshift=2.8]crops raw.east) -- (Feature Extractor);
  \draw[connector] (Feature Extractor) -- (FEPCD Align);
  \draw[connector] (FEPCD Align) -- (Fusion);
  \draw[connector] (Fusion) -- (LRCN);

  \draw[rectangle connector=0cm] (crops raw.north) to (circle.north);

  \draw[connector] (LRCN) -- (circle);
  \draw[connector] (circle) -- ([yshift=2.8]final.west);

  \draw[dashed] (0.50, -0.4) -- (0.50, +0.4);
  \draw[dashed] (1.70, -0.4) -- (1.70, +0.4);
  \draw[dashed] (2.26, -0.4) -- (2.26, +0.4);
  \draw[dashed] (3.25, -0.4) -- (3.25, +0.4);
 
\end{tikzpicture}%